\documentclass[a4paper,fleqn]{cas-dc}
\ExplSyntaxOn
\cs_gset:Npn \__first_footerline:
  { \group_begin: \small \sffamily \__short_authors: \group_end: }
\ExplSyntaxOff

\usepackage{svg}
\usepackage[numbers]{natbib}
\usepackage{float}

\def\tsc#1{\csdef{#1}{\textsc{\lowercase{#1}}\xspace}}
\tsc{WGM}
\tsc{QE}

\begin{document}
\let\WriteBookmarks\relax
\def\floatpagepagefraction{1}
\def\textpagefraction{.001}
\shortauthors{Guido Manni et~al.}

\title [mode = title]{BodySLAM: A Generalized Monocular Visual SLAM Framework for Surgical Applications}  
\shorttitle{}

\author[1,2]{Guido Manni}

\credit{Conceptualization, Methodology, Software, Validation, Formal analysis, Investigation, Writing - Original Draft, Writing - Review \& Editing, Visualization}

\ead{guido.manni@unicampus.it}

\author[2]{Clemente Lauretti}
\credit{Conceptualization, Methodology, Formal analysis, Writing - Review \& Editing, Supervision}

\ead{c.lauretti@unicampus.it}

\author[3]{Francesco Prata}
\credit{Resources, Data Curation}
\ead{f.prata@policlinicocampus.it}

\author[3]{Rocco Papalia}
\credit{Resources, Data Curation}
\ead{rocco.papalia@policlinicocampus.it}

\author[2]{Loredana Zollo}
\credit{Conceptualization, Methodology, Writing - Review \& Editing, Supervision}
\ead{l.zollo@unicampus.it}

\author[1]{Paolo Soda}
\credit{Conceptualization, Methodology, Formal analysis, Writing - Review \& Editing, Supervision, Project administration, Funding acquisition}
\ead{p.soda@unicampus.it}





\affiliation[1]{organization={Research Unit of Computer Systems and Bioinformatics, Department of Engineering, Università Campus Bio-Medico di Roma, Rome, Italy}}

\affiliation[2]{organization={Unit of Advanced Robotics and Human-Centred Technologies, Department of Engineering, Università Campus Bio-Medico di Roma, Rome, Italy}}

\affiliation[3]{organization={Department of Urology, Fondazione Policlinico Universitario Campus Bio-Medico, Rome, Italy}}

%

\begin{abstract}
Endoscopic surgery relies on two-dimensional views, posing challenges for surgeons in depth perception and instrument manipulation.
While Monocular Visual Simultaneous Localization and Mapping (MVSLAM) has emerged as a promising solution, its implementation in endoscopic procedures faces significant challenges due to hardware limitations, such as the use of a monocular camera and the absence of odometry sensors.
This study presents BodySLAM, a robust deep learning-based MVSLAM approach that addresses these challenges through three key components: CycleVO, a novel unsupervised monocular pose estimation module; the integration of the state-of-the-art Zoe architecture for monocular depth estimation; and a 3D reconstruction module  creating a coherent surgical map. 
The approach is rigorously evaluated using three publicly available datasets (Hamlyn, EndoSLAM, and SCARED) spanning laparoscopy, gastroscopy, and colonoscopy scenarios, and benchmarked against four state-of-the-art methods. 
Results demonstrate that CycleVO exhibited competitive performance with the lowest inference time among pose estimation methods, while maintaining robust generalization capabilities, whereas Zoe significantly outperformed existing algorithms for  depth estimation in endoscopy.
BodySLAM's strong performance across diverse endoscopic scenarios demonstrates its potential as a viable MVSLAM solution for endoscopic applications.
\end{abstract}



\begin{keywords}
Endoscopic SLAM \sep 
Monocular depth estimation \sep
Monocular Pose estimation \sep 
Deep learning \sep
Surgical navigation\sep
\end{keywords}

\maketitle

\section{Introduction}\label{Int}
Laparoscopy has become the preferred surgical method for various procedures across medical specialties due to its minimally invasive nature \cite{SCHMIDT2024103131}.
However, a significant limitation of this technique is its reliance on two-dimensional views \cite{DURRANI1995237}, and inferring depth from 2D images can be challenging for surgeons and requires years of experience \cite{Stewart2007ThePO}. \\
To address this challenge, there has been growing interest in integrating 3D imaging technology into surgical procedures. 
Studies have shown that 3D imaging reduces cognitive strain on surgeons, enhances procedural performance, and increases the accuracy of instrument manipulation \cite{Smith2014EffectOP, Vettoretto2018WhyLM}. 
These benefits lead to better procedural outcomes and shortened learning curves for surgical trainees \cite{Srensen2015ThreedimensionalVT, Davies2020ThreedimensionalVT, Schwab2017EvolutionOS}. \\
In recent years, camera-based tracking and mapping methods have gained popularity in surgical 3D imaging research~\cite{SCHMIDT2024103131}. 
Approaches such as mosaicking and Shape from Template have been explored, but each faces specific limitations in the surgical context.
Mosaicking, which stitches together overlapping images to create a comprehensive view of the scene \cite{Bergen2016StitchingAS, Bano2022PlacentalVH, Li2021GloballyOF}, struggles with illumination variations, camera motion, moving objects, and noise, which can lead to artifacts and reduced visual quality in surgical environments.
Shape from Template estimates the 3D shape of a deformable object by aligning a known template to observed images \cite{Cheema2019ImageAlignedDL, Zhang20213DRO}, but, it is affected by template size and complexity, image quality and noise, rotation and scaling issues, partial occlusion, and the presence of similar shapes or clutter in surgical scenes. \\
In contrast, Simultaneous Localization and Mapping (SLAM) has emerged as a promising solution to address these limitations. 

SLAM algorithms simultaneously estimate the camera pose and construct a map of the environment in real-time, so that they can effectively handle the dynamic and complex nature of surgical scenes, including moving objects, occlusions, and deformations.
However, implementing SLAM in endoscopic procedures presents significant challenges due to hardware limitations and environmental factors.
The use of monocular cameras, necessitated by size constraints, and the absence of odometry sensors force SLAM to rely solely on visual information for localization and mapping. 
This approach is known as Monocular Visual SLAM (MVSLAM).
This reliance on visual data is further complicated by the unique characteristics of surgical images, which often lack discernible feature \cite{Widya2019WholeS3} and exhibit lighting variations \cite{Makki2023EllipticalSD, Zhang2022SLAMTKARI}. 
These factors make robust feature extraction and matching particularly difficult for traditional feature-based MVSLAM methods, such as ORB-SLAM \cite{ORBSLAM3_TRO} and Structure from Motion \cite{Longuet-Higgins1981}.
Additional complications arise from organ deformation \cite{Schle2022AMS}, and the presence of blood, fluids, and smoke \cite{Li2019SuPerAS,Liu2023SurfaceDT}. 
These environmental challenges significantly impair the performance of traditional MVSLAM approaches in endoscopic settings. \\ 
Recent works \cite{Recasens2021EndoDepthandMotionRA, Ozyoruk2021EndoSLAMDA, shao2022self, Teufel2024} have explored deep learning-based approaches to address these challenges. 
However, these methods, while promising, still struggle with the same environmental factors that plague traditional approaches. 
Furthermore, these approaches face additional challenges in generalizing to unseen surgical environments due to data scarcity and the wide variability of surgical scenes.

To overcome these limitations in this paper we propose a robust deep learning MVSLAM approach that effectively works in diverse surgical settings, referred to as {\em BodySLAM} in the following.  
It introduces  a newly developed model and  state-of-the-art   that, together, overcome the issues of low-texture surfaces, lighting variations, and generalization to unseen datasets. 
Our key contributions are: 

\begin{itemize}
\item CycleVO: a novel GAN-based pose estimation module that enhances accuracy and robustness in surgical settings while addressing data scarcity. 
This unsupervised method uniquely applies concepts from CycleGAN and InfoGAN to learn robust pose representations from unlabelled endoscopic video sequences.

\item Integration of the Zoe model \cite{zoe} for monocular depth estimation: we leverage its superior generalization capabilities across diverse datasets without fine-tuning, marking the first use of such a well-generalized model for depth prediction in endoscopic MVSLAM. 
We present an in-depth validation procedure demonstrating its robust performance across various endoscopic scenarios.

\end{itemize}

To rigorously test our approach and demonstrate its generalization capabilities, we conducted comprehensive evaluations using the largest publicly available endoscopic MVSLAM dataset to date. Our experiments span three surgical environments: laparoscopy, gastroscopy, and colonoscopy.
We utilized the Hamlyn \cite{Recasens2021EndoDepthandMotionRA}, EndoSLAM \cite{Ozyoruk2021EndoSLAMDA}, and \\ SCARED~\cite{EndoVis2019} datasets, each offering unique features for validating our proposed method. 
We also benchmarked our approach against four state-of-the-art methods \cite{Recasens2021EndoDepthandMotionRA, Ozyoruk2021EndoSLAMDA, shao2022self, Teufel2024},  which are overviewed in section \ref{subsec:DL-based-MVSLAM} and then presented in section \ref{sec:ExC}.
The remainder of this paper is organized as follows: section \ref{sec:RL} provides an overview of the related work. 
Section~\ref{sec:Methods} describes our proposed approach, detailing the three primary modules of our framework.
Section \ref{sec:Mat} presents the materials used in this study, including the training, validation, and testing datasets.
Section \ref{sec:ExC} outlines the experimental configuration, including the comparative analysis against the two state-of-the-art approaches, performance metrics, and statistical analysis. 
Section \ref{sec:Res} presents the experimental results. 
Finally, section \ref{sec:Disc&ftw} provides concluding remarks.

\section{Related Works}
\label{sec:RL}

\subsection{\textbf{Traditional Monocular Visual SLAM Architecture}}
\begin{figure}
	\centering
	\includegraphics[width=0.50\textwidth]{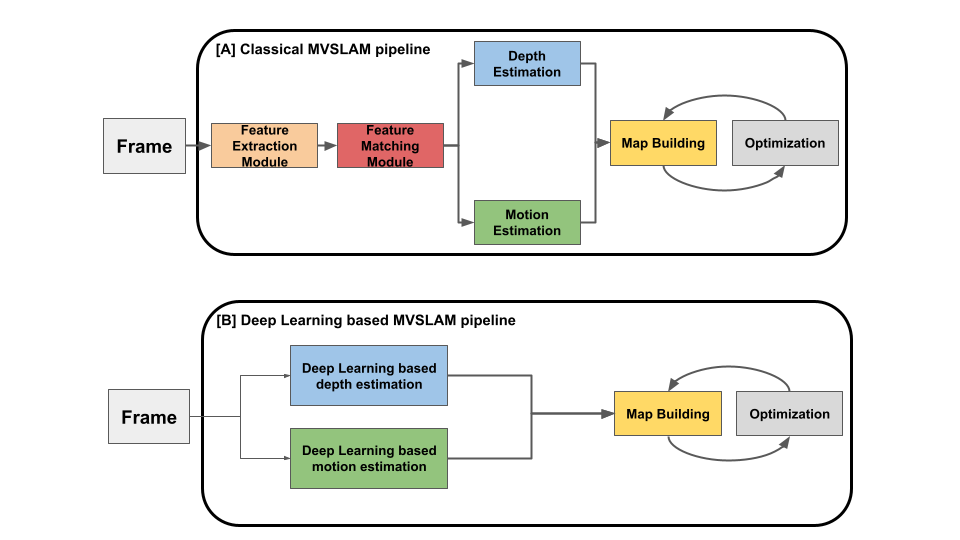}
	\caption{Comparison of (A) a classical MVSLAM system and (B) a fully deep learning-based MVSLAM framework. The traditional approach consists of feature extraction, feature matching, motion estimation, depth map estimation, map building, and optimization components. In contrast, the deep learning-based framework replaces conventional algorithms with deep learning models for feature extraction, feature matching, motion estimation, and depth estimation, potentially simplifying the overall architecture complexity.}
	\label{fig:typical_MVSLAM}
\end{figure}

A classical Monocular Visual SLAM (MVSLAM) system typically consists of several integrated components, as illustrated in Figure \ref{fig:typical_MVSLAM}A. The framework begins with a feature extraction module, which identifies salient features (like ORB, SURF, and SIFT \cite{Rublee2011ORBAE, Lowe2004DistinctiveIF, Bay2006SURFSU}) within the images.
A feature-matching algorithm then establishes correspondences between these features across consecutive frames.
The matched features are passed into a motion estimation algorithm and a depth map estimation algorithm to estimate the relative motion between consecutive frames and generate a sparse depth map, respectively.
These outputs are utilized by a map-building algorithm to reconstruct the environmental model.
Additionally, the MVSLAM framework incorporates an optimization module, which executes procedures such as bundle adjustment \cite{Triggs1999BundleA} and pose graph optimization to refine the map and pose estimates.
\subsection{\textbf{Challenges and Limitations of Traditional MVSLAM in Endoscopic Surgery}}
Traditional feature-based methodologies in MVSLAM have encountered notable challenges in the context of endoscopic surgery due to the unique nature of surgical images.
Endoscopic images often lack discernible features and are subject to lighting variations, posing substantial difficulties for feature extraction and matching algorithms.
For example, a modified version of ORB-SLAM \cite{Mahmoud}, which utilized a limited number of keyframes for dense reconstruction, struggled with soft, texture-less tissue, a common characteristic in minimally invasive surgery.
This highlights the inherent challenge of detecting robust features in environments where texture and contrast are minimal.

Early SLAM techniques, such as those relying on Structure from Motion \cite{Zhao2016TheEA}, aimed to concurrently track camera motion and reconstruct the 3D environment.
However, these methods were severely limited by the low-texture environments typical of minimal invasive surgert.
Structure from Motion relies on robust wide-baseline feature matching, which becomes untenable in texture-deficient surgical scenes.
Moreover, the computational expense of Structure from Motion is another drawback, as it requires processing all images collectively to optimize the 3D structures and camera poses, which is not always feasible in real-time surgical applications.

\subsection{\textbf{Recent Advancements in Deep Learning-based Endoscopic MVSLAM}}
\label{subsec:DL-based-MVSLAM}
To address the limitations of traditional MVSLAM methods in endoscopic surgery, researchers have turned to deep learning-based approaches. As shown in Figure \ref{fig:typical_MVSLAM}B, a fully deep learning-based MVSLAM framework replaces the conventional feature matching, feature extraction, motion estimation, and depth estimation algorithms with deep learning models. These models can learn to extract relevant features and estimate depth and motion directly from the input images, even in the presence of low-texture environments and lighting variations.

Deep learning models offer several advantages over traditional MVSLAM methods in the context of endoscopic surgery. They can adapt to the dynamic, texture-sparse, and unpredictable nature of surgical environments by learning to extract relevant features and estimate depth and motion directly from the data. This eliminates the need for hand-crafted feature detectors and matching algorithms, which often struggle in low-texture environments.

Recent years have witnessed significant advancements in deep learning-based MVSLAM for endoscopic surgery. 
EndoSLAM \cite{Ozyoruk2021EndoSLAMDA} represents a fully end-to-end approach, leveraging convolutional neural networks and attention mechanisms for monocular depth and pose estimation in gastroscopy. 
This work demonstrated the feasibility of deep learning solutions for domain-specific endoscopic SLAM.
Hybrid approaches have also emerged, combining deep learning with traditional techniques. 
Recasens et al. \cite{Recasens2021EndoDepthandMotionRA} proposed the Endo-Depth-and-Motion framework, which integrates MonoDepthV2 \cite{monodepth2} for depth estimation with a PTAM-inspired \cite{Klein2007ParallelTA} photometric method for pose estimation. 
While this approach showed improvements over purely traditional methods, it still encountered limitations in extended procedures.
Addressing a critical challenge in endoscopic environments, Shao et al. \cite{shao2022self} introduced the concept of appearance flow to handle brightness inconsistency. 
Their self-supervised framework employs a generalized dynamic image constraint, relaxing the conventional brightness constancy assumption and enhancing robustness to varying illumination conditions.
Most recently, Teufel et al. \cite{Teufel2024} presented OneSLAM, a generalized monocular SLAM approach for endoscopy utilizing tracking any point (TAP) foundation models. 
This method achieves sparse correspondences across multiple frames for joint optimization of camera poses and anatomical geometry. Notably, OneSLAM demonstrates zero-shot generalization across various endoscopic domains without retraining, while maintaining or surpassing the performance of domain-specific approaches.
These advancements illustrate the growing capabilities of deep learning in addressing the unique challenges of endoscopic SLAM, from domain-specific solutions to more generalized approaches that promise broader applicability in surgical environments.

\subsection{\textbf{Challenges in Deep Learning-based MVSLAM for Endoscopic Surgery}}
Despite recent advancements, deep learning-based MVSLAM for endoscopic surgery faces several challenges. The complex surgical environment presents unique obstacles that must be overcome \cite{Schle2022AMS, Li2019SuPerAS,Liu2023SurfaceDT}. 
A major hurdle is the requirement for large, diverse datasets to train generalizable models, which are costly and time-consuming to collect and annotate. 
Additionally, the computational demands of deep learning models can impede real-time performance in resource-limited surgical settings. 
Striking a balance between model accuracy and efficiency remains crucial for practical implementation.

\section{Methods}\label{sec:Methods}
\begin{figure*}[htbp]
	\centering
	\includegraphics[width=0.9\textwidth]{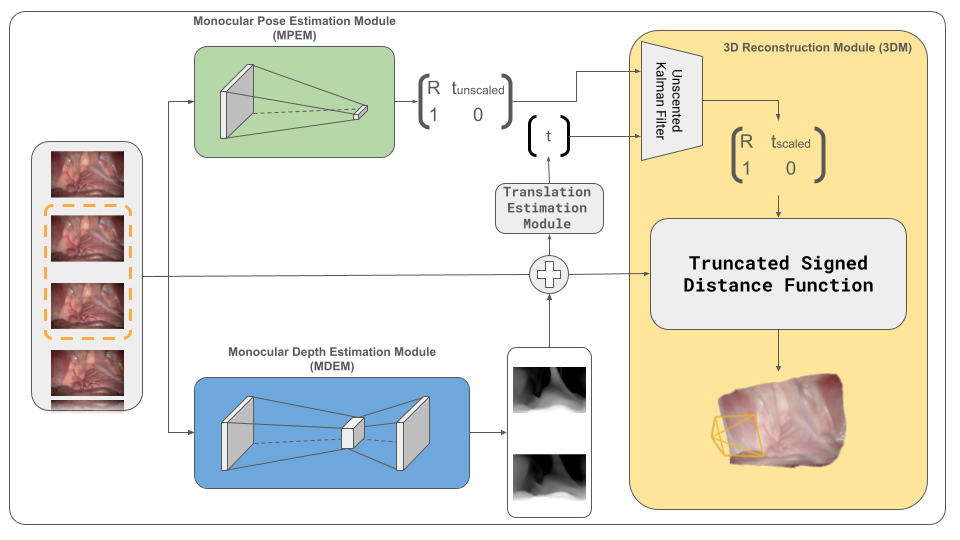}
	\caption{The approach takes RGB frames as input and outputs a 3D reconstruction of the surgical scene. The MPEM, utilizing CycleVO, computes relative motion between consecutive frames, outputting a motion matrix $M = [R, t_{unscaled}, 1, 0]$, where $R$ is the rotation matrix and $t_{unscaled}$ is the unscaled translation vector. The MDEM estimates depth maps from RGB frames. The Translation Estimation Module estimates a scaled translation vector $t_{scaled}$, which is then combined with $t_{unscaled}$ using an Unscented Kalman Filter to correct the scale of the motion matrix. Finally, the 3DM combines the RGB frames, depth maps, and scaled pose matrices to generate and update the 3D model}
	\label{fig:BoySLAM_Pipeline}
\end{figure*}
The proposed MVSLAM approach is composed of three key modules (\figurename~\ref{fig:BoySLAM_Pipeline}):  the Monocular Pose Estimation Module (MPEM), the Monocular Depth Estimation Module (MDEM), and the 3D Reconstruction Module (3DM). 
The structure of the framework is designed to ensure that, while the  MPEM and the MDEM operate independently, their outputs synergistically contribute to the effective functioning of the 3DM.
The MPEM (\sectionautorefname~\ref{subsec:MPEM}) processes  the RGB input using our novel unsupervised deep network, named as CycleVO, to estimate the relative motion of the camera between consecutive frames, outputting a motion matrix $M = [R, t_{unscaled}, 1, 0]$, where $R$ is the rotation matrix and $t_{unscaled}$ is the unscaled translation vector due to the scale ambiguity inherent in monocular pose estimation.
Simultaneously, the MDEM (\sectionautorefname~\ref{subsec:MDEM}) estimates the depth map of the scene from the RGB input.
Given that the translation component ($t_{unscaled}$) of the relative motion matrix is unscaled, a dedicated module, comprising a translation estimation module and an Unscented Kalman Filter, is tasked with scale correction. 
The translation estimation module estimates a scaled translation vector $t_{scaled}$, which is then combined with $t_{unscaled}$ using the UKF to correct the scale of the motion matrix.
The 3DM (\sectionautorefname~\ref{subsec:3DM}) then employs the outputs from both the MDEM and MPEM. 
It utilizes the depth map from the MDEM and the pose estimations from the MPEM to generate an initial point cloud of the current field of view of the endoscopic camera. This point cloud is carefully merged with the broader scene point cloud, leveraging the pose information to ensure precise alignment and integration.

\subsection{Monocular Pose Estimation Module (MPEM): CycleVO}
\label{subsec:MPEM}
\begin{figure*}[htbp]
\centering
\includegraphics[width=0.9\textwidth]{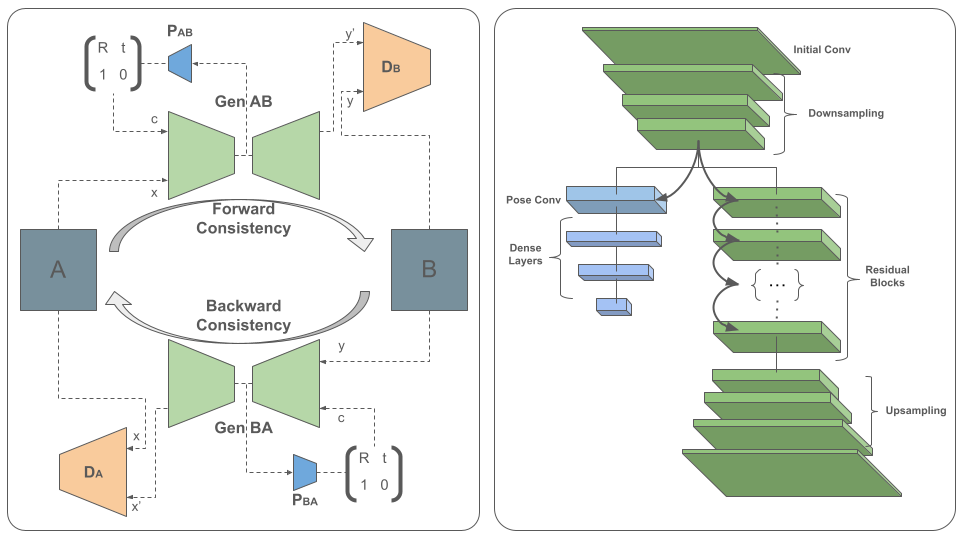}
	\caption{Integration of Pose Estimation within the Cycle Consistency Framework and the Neural Network Architecture. \textit{Left:} The diagram illustrates the integration of pose estimation within the Cycle Consistency framework. Generators $\text{Gen}{AB}$ and $\text{Gen}{BA}$ perform domain transformations, while pose networks $P_{AB}$ and $P_{BA}$ predict the relative pose between consecutive frames. The predicted pose $M$ is concatenated with the latent space $z$ to improve image-to-image translation performance.
\textit{Right:} The neural network architecture for pose estimation includes convolutional, downsampling, residual, and upsampling layers. The bottleneck is modified to accommodate pose estimation, where the generator encoder $E$ processes concatenated frames $f_c = [f_{i-1}, f_i]$ and the pose estimation tail $P$ produces the relative motion matrix $M = [R, t_{unscaled}, 1, 0]$, where $R$ is the rotation matrix and $t_{unscaled}$ is the unscaled translation vector. The latent space $z$ and predicted pose $M$ are concatenated and fed into the generator $G$ to produce the next frame $\hat{f}_i$.}
\label{fig:pose_architecture}
\end{figure*}

The MPEM introduces {\em CycleVO}, a novel unsupervised method designed to address the challenge of data scarcity in endoscopic environments. While CycleVO incorporates architectural ideas insipired by CycleGAN \cite{cyclegan} for cyclic image translation and concepts from InfoGAN \cite{Chen2016InfoGANIR} for latent space manipulation, it uniquely applies and extends these concepts to pose estimation.
The key innovation lies in learning the estimation of relative camera pose between consecutive frames through the pre-task of image-to-image translation.
We define this pre-task as a function $G: f_{i-1} \rightarrow f_i$, which generates the current frame $f_i$ from the previous frame $f_{i-1}$. Similarly, we define the function $F: f_i \rightarrow f_{i-1}$, which generates the previous frame $f_{i-1}$ from the current frame $f_i$.
To improve the performance of this image-to-image translation pre-task, we draw inspiration from InfoGAN's manipulation of the latent space through the maximization of mutual information.
We propose to manipulate the latent space $z$ produced by the generator encoder $E$ by concatenating a prediction of the relative pose $M$ between the two consecutive frames to $z$.
The underlying concept is that if the predicted pose $M$ is correct, the image-to-image translation task $G(f_{i-1}, M) \approx f_i$ will also be successful.
Conversely, if the estimated pose is incorrect, the pre-task will fail.
This novel approach creates a synergy between pose estimation and image generation, enabling the model to learn robust pose representations from unlabeled endoscopic video sequences.
The CycleVO generator performs two key tasks (right panel of \figurename~\ref{fig:pose_architecture}):
\begin{itemize}
\item Predicting the pose $M$ between two concatenated consecutive frames ($f_{i-1}$ and $f_i$). This is done by passing the concatenated frame $f_c = [f_{i-1}, f_i]$ through the generator encoder $E$, which then feeds into the pose estimation tail $P$. The pose network $P$ outputs the relative motion matrix $M$ in quaternion form to avoid gimbal lock and improve numerical stability. The quaternion is immediately converted to a matrix representation $M = [R, t_{unscaled}, 1, 0]$, where $R$ is the rotation matrix and $t_{unscaled}$ is the unscaled translation vector, for better integration within the SLAM pipeline. The final linear layer of the pose network is responsible for outputting $M$.
\item Generating the next frame (image-to-image translation pre-task).
First, the previous frame $f_{i-1}$ is passed through the encoder $E$. 
The resulting latent space $z = E(f_{i-1})$ is then concatenated with the predicted pose $M$ from the first task. 
This concatenated latent space $[z, M]$ is then fed into the generator $G$, which produces the next frame $\hat{F}i = G(z, M)$.
\end{itemize}
We now discuss the objective loss function, which has been optimized for the task of pose estimation.
The objective function consists of three terms:
\begin{itemize}
\item The adversarial loss for matching the distribution of generated images to the data distribution in the target domain.
\item The cycle consistency loss to prevent the learned mappings on frames $G$ and $F$ from contradicting each other.
\item The cycle consistency loss on pose to prevent the learned mappings from contradicting each other regarding the estimated pose.
\end{itemize}
The full objective loss is defined as follows:
\begin{equation}
\begin{split}
\mathcal{L}(G, F, D_X, D_Y) = \mathcal{L}{_\text{adv}}(G, D_Y, X, Y) \\ 
+\mathcal{L}{_\text{adv}}(F, D_X, Y, X) + \lambda_1 \mathcal{L}{_\text{cycImg}}(G, F) \\
+\lambda_2\mathcal{L}{_\text{cycPose}}(G, F)
\end{split}
\label{eqn: cycle_gan_full_loss}
\end{equation}
where the first two terms, $\mathcal{L}{_\text{adv}}(G, D_Y, X, Y)$ and \\ $\mathcal{L}{_\text{adv}}(F, D_X, Y, X)$, are the adversarial losses.
The third term $\lambda_1\mathcal{L}{_\text{cycImg}}(G, F)$ is the classical cycle consistency loss on the generated images.
The final term $\lambda_2\mathcal{L}{_\text{cycPose}}(G, F)$ is a custom term added to enforce pose consistency between the estimated poses from the generated images.
To enforce pose consistency, the term $\mathcal{L}{_\text{cycPose}}(G, F)$ is defined as follows:
\begin{equation}
\begin{split}
\mathcal{L}{_\text{cycPose}}(G, F) = \mathcal{L}{_\text{chordal}}(R_{i-1,i}, \hat{R}_{i-1,i}) \\
+ \mathcal{L}_1(t_{i-1,i}, \hat{t}_{i-1,i}) + \mathcal{L}{_\text{chordal}}(R_{i,i-1}, \hat{R}_{i,i-1}) \\
+ \mathcal{L}_1(t_{i,i-1}, \hat{t}_{i,i-1})
\end{split}
\end{equation}
where $R_{i-1,i}$ and $\hat{R}_{i-1,i}$ are the rotation components of the estimated pose between the real frames ($f_{(i-1)real}$, $f_{(i)real}$) and the generated frames ($f_{(i-1)recov}$, ${F}_{(i)real}$), respectively. 
Similarly, $t_{i-1,i}$ and $\hat{t}_{i-1,i}$ are the translation vectors of the estimated pose between the real frames ($f_{(i-1)real}$, $f_{(i)real}$) and the generated frames ($f_{(i-1)recov}$, $f_{(i)real}$), respectively. The terms $R_{i,i-1}$, $\hat{R}_{i,i-1}$, $t_{i,i-1}$, and $\hat{t}_{i,i-1}$ are defined analogously for the mapping $f$.
The chordal loss $\mathcal{L}{_\text{chordal}}$ is defined as:
\begin{equation}
\mathcal{L}_{\text{chordal}}(R, \hat{R}) = |R - \hat{R}|_F
\end{equation}
where $|\cdot|_F$ denotes the Frobenius norm. The $\mathcal{L}1$ loss is defined as:
\begin{equation}
\mathcal{L}_1(t, \hat{t}) = \sum{i=1}^{n} |t_i - \hat{t}i|
\end{equation}
where $n$ is the dimension of the translation vector. The chordal loss constrains the rotation component of the estimated relative pose, while the $\mathcal{L}1$ loss constrains the translation component. \\
The overall workflow (left part of \figurename~\ref{fig:pose_architecture}) is as follows:
\begin{itemize}
\item The concatenated frame $f_c = [F{i-1}, f_i]$ is passed through the network to predict the relative pose $M = P(E(F_c))$.
\item The previous frame $f{i-1}$ is then passed through the encoder $E$ for the image-to-image translation task, producing the latent space $z = E(F{i-1})$.
\item At the encoder level, the predicted pose $M$ is concatenated with the latent space $z$, resulting in $[z, M]$.
\item The concatenated latent space $[z, M]$ is then fed into the generator $G$, which produces the next frame $\hat{F}_i = G(z, M)$.
\end{itemize}

\subsection{Monocular Depth Estimation Module (MDEM): Zoe}
\label{subsec:MDEM}
The Monocular Depth Estimation Module is a crucial component of our MVSLAM approach, responsible for estimating depth from the endoscopic camera frames.
Accurate depth estimation is essential for the subsequent 3D reconstruction process. \\
To address the challenges of monocular depth estimation in endoscopic environments, particularly the issue of data scarcity, we employ the innovative Zoe architecture \cite{zoe}. This represents the first use of a well-generalized model for depth prediction in endoscopic SLAM, tackling the data scarcity problem inherent in medical imaging applications. \\
Zoe model utilizes a distinctive two-stage framework:

\begin{enumerate}
    \item Pre-training on a wide array of datasets for relative depth estimation, which fosters excellent generalization ability. The datasets used for pre-training include.
    \item Adding heads for metrics depth estimation and fine-tuning on metrics depth datasets.
\end{enumerate}

This approach allows Zoe to maintain metrics scale while benefiting from the generalization capabilities obtained during the relative depth pre-training phase.\\
Zoe architecture features a novel metrics bins module, which takes multi-scale features from the MiDaS \cite{Ranftl} decoder and predicts bin centres crucial for metrics depth prediction. 
The module predicts all bin centres at the initial stage and adjusts them at subsequent decoder layers using attractor layers, enabling more accurate and adaptable depth estimation. \\
Another innovative aspect of Zoe is its use of the log-binomial method for final metrics depth prediction, instead of the conventional softmax approach. 
This method linearly combines the bin centres, weighted by their probability scores, enhancing the model's ability to accurately predict depth in a structured and ordered manner. \\
Integrating Zoe into our MVSLAM approach represents a significant advancement in-depth estimation for endoscopic applications. 
Its architecture, combining the strengths of relative and metrics depth estimation and utilizing novel components, makes it exceptionally suited for the challenging and variable conditions of endoscopic surgery. \\
In our approach, we chose to assess Zoe's out-of-the-box performance to evaluate its generalization capabilities across domains, without retraining or fine-tuning the model specifically for the surgical context (\sectionautorefname~\ref{subsec:MDEM-Res}).

\subsection{3D Reconstruction Module (3DM)}
\label{subsec:3DM}
The 3D reconstruction module generates three-dimensional models from endoscopic images through a multi-phase process. The key steps involved are:
\begin{enumerate}
    \item Constructing a point cloud from the endoscopic camera perspective using pseudo-RGBD frames produced by the MDEM.
    \item Refining the pose estimate provided by the MPEM through multiple optimization steps.
    \item Aligning and integrating the newly acquired point cloud with the pre-existing reconstructed scene, guided by the refined pose estimate.
\end{enumerate}
One of the critical aspects of this module is the optimization procedures applied to enhance the pose estimate derived from the MPEM. 
The raw pose estimate cannot be directly employed for point cloud alignment and merging due to scale ambiguity inherent in monocular pose estimation. \\
To address this issue, a dedicated sub-module for scale correction has been implemented. This sub-module employs a classical pose estimation algorithm that leverages features extracted (SIFT or ORB) from pseudo-RGBD frames to estimate a pose. Although less effective for pose estimation in low-texture environments, it is viable for deriving a pseudo-scale. \\
An Unscented Kalman Filter (UKF) is utilized to fuse the unscaled translation vector $t_{unscaled}$ obtained from the MPEM with the scaled translation vector $t_{scaled}$ estimated by the translation estimation module. The state vector of the UKF is defined as:
\begin{equation}
x = [t_x, t_y, t_z]^T
\end{equation}
where $t_x$, $t_y$, and $t_z$ are the components of the translation vector. The measurement model of the UKF is given by:
\begin{equation}
z = [t_{scaled_x}, t_{scaled_y}, t_{scaled_z}]^T
\end{equation}
where $t_{scaled_x}$, $t_{scaled_y}$, and $t_{scaled_z}$ are the components of the scaled translation vector obtained from the RGB-D odometry or the scaling factor computation.

The UKF operates in two main steps:
\begin{enumerate}
    \item prediction step: The UKF predicts the next state $\hat{x}$ using the unscaled translation vector $t_{unscaled}$ obtained from the MPEM:
    \begin{equation}
    \hat{x} = f(x, t_{unscaled})
    \end{equation}
    where $f$ is the state transition function that uses the MPEM to predict the next state.
    
    \item Update step: The UKF updates the predicted state $\hat{x}$ using the scaled translation vector $t_{scaled}$ obtained from the RGB-D odometry or the scaling factor computation:
    \begin{equation}
    x = \hat{x} + K(z - h(\hat{x}))
    \end{equation}
    where $K$ is the Kalman gain, $z$ is the measurement vector containing the scaled translation, and $h$ is the measurement function that maps the predicted state to the measurement space.
\end{enumerate}

The corrected translation vector $t_{corrected}$ is then obtained from the updated state of the UKF:
\begin{equation}
t_{corrected} = [x_1, x_2, x_3]^T
\end{equation}
where $x_1$, $x_2$, and $x_3$ are the scaled translation components of the updated state vector.

The corrected translation vector $t_{corrected}$ is used to update the motion matrix $M$ obtained from the MPEM, resulting in a scaled motion matrix $M_{scaled}$:
\begin{equation}
M_{scaled} = [R, t_{corrected}, 1, 0]
\end{equation}
where $R$ is the rotation matrix obtained from the MPEM. \\
Additionally, pose graph optimization is periodically employed to mitigate the cumulative effect of drift by evenly distributing the error across all estimated poses. \\
Upon aligning the point cloud, the module utilizes the Truncated Signed Distance Function (TSDF) \cite{Curless1996AVM} technique to convert the point cloud data into a comprehensive volumetric representation. The TSDF is a volumetric representation that stores the signed distance to the nearest surface at each voxel. It allows for efficient fusion of depth information from multiple viewpoints and enables the generation of a smooth and continuous surface reconstruction.
The scene is partitioned into voxels, each storing a cumulative signed distance function. 
This voxel-based approach is continually updated through sequential averaging, reflecting the most current information captured by the endoscopic camera. 

\section{Materials}
\label{sec:Mat}

In this study, we employed a comprehensive approach to train, validate, and test our proposed method, utilizing the diverse datasets listed in table \ref{table:datasets} to ensure robustness and generalizability in endoscopic applications.

\begin{table}[htbp]
\centering
\footnotesize
\setlength{\tabcolsep}{3pt}
\begin{tabular}{@{}lcccr@{}}
\toprule
\textbf{Dataset} & \textbf{Depth} & \textbf{Pose} & \textbf{Procedure} & \textbf{Frames} \\
\midrule
Hamlyn \cite{Recasens2021EndoDepthandMotionRA} & $\checkmark$ & & Laparoscopy & 78,160 \\
EndoSLAM \cite{Ozyoruk2021EndoSLAMDA} & & $\checkmark$ & Gastroscopy, Colonoscopy & 76,837 \\
SCARED \cite{EndoVis2019} & $\checkmark$ & $\checkmark$ & Laparoscopy & 17,206 \\
\bottomrule
\end{tabular}
\caption{Overview of endoscopic datasets used for evaluating depth estimation and pose estimation modules in MVSLAM framework. Checkmarks indicate the specific modules tested on each dataset}
\label{table:datasets}
\end{table}

\subsection{\textbf{Training Datasets for CycleVO}}
The training and validation processes were performed on the CycleVO module because it was the only model we trained in this study. 
Two distinct datasets were utilized: an extensive internal dataset for training and selected subsets from the EndoSLAM dataset \cite{Ozyoruk2021EndoSLAMDA} for validation.
For training, we leveraged a large internal unlabeled dataset comprising more than 300 hours of gastroscopy and prostatectomy videos, collected from 100 patients undergoing these procedures at Fondazione Policlinico Universitario Campus Bio-Medico. In total, this training dataset contained 2,250,900 frames, allowing us to effectively train our pose estimation network on a substantial amount of endoscopic data. 
To prepare the training data, we processed the videos by extracting individual frames and performing a center crop of size 128x128 pixels on each frame. This center cropping approach enhances system robustness by training it to work effectively with a reduced receptive field, while avoiding any distortion that may arise from resizing the frames. 
By focusing on the central region of each frame, we ensure that the most relevant visual information is preserved while maintaining the spatial integrity of the data.
To validate CycleVO, we used the following subsets from the EndoSLAM dataset, which are completely separate from the training data: HighCam Colon IV Tumor-Free Trajectory 1, LowCam Colon IV Tumor-Free Trajectory 1, HighCam Colon IV Tumor-Free Trajectory 5, and LowCam Colon IV Tumor-Free Trajectory 5. These subsets account for four video sequences with a total of 3024 frames, providing a diverse range of scenarios to assess CycleVO's inference performance. Importantly, there is no overlap between the training and validation datasets, as the EndoSLAM subsets were specifically chosen for validation purposes only and were not used in any way during the training phase.

\subsection{\textbf{Testing Datasets for Depth and Pose Estimation Modules}}
To evaluate the performance of our depth estimation and pose estimation modules we utilized three comprehensive datasets  that represent the largest publicly available endoscopic MVSLAM datasets to date.
 These datasets, summarized in \tablename~\ref{table:datasets}, provide diverse scenarios and data type to assess the robustness and generalizability of our method:
\begin{enumerate}
    \item Hamlyn Dataset \cite{Recasens2021EndoDepthandMotionRA}: We employed an enhanced version of the Hamlyn dataset, which contains stereo images from endoscopic interventions on porcines. 
    This dataset, augmented with depth ground truth derived from stereo images using Libelas, consists of 78,160 frames across 20 videos, making it suitable for testing the depth estimation module of our method. 
    However, it lacks pose ground truth, limiting its utility for assessing the pose estimation module.
    \item EndoSLAM Dataset \cite{Ozyoruk2021EndoSLAMDA}: In addition to its use in validating CycleVO, the EndoSLAM dataset was also employed for testing the pose estimation module of our method.
    The subsets used for validating CycleVO were excluded from the testing process to maintain a clear separation between validation and testing data.
    While the EndoSLAM dataset includes depth data, we did not use it for testing depth estimation due to its synthetic nature.
    The dataset comprises both ex-vivo and synthetic porcine data, with 35 sub-datasets totaling 76,837 frames spanning various gastrointestinal areas, providing pose data for a comprehensive evaluation of our method's performance in pose estimation.
    \item SCARED Dataset \cite{EndoVis2019}: The SCARED dataset, recorded using a da Vinci Xi surgical robot, consists of 7 training and 2 test sets, each corresponding to a single porcine subject. 
    Each set contains 4 to 5 keyframes representing distinct scene views, with structured light patterns projected for dense stereo reconstruction. 
    The movement of the endoscope enabled the capture of camera poses relative to the robot base, and the projection of 10-bit Gray code patterns onto the scene ensured unique pixel encoding for efficient stereo matching and depth recovery. 
    The SCARED dataset provides both depth maps and ground truth poses, making it suitable for testing both the depth and pose estimation modules of our method.
\end{enumerate}
By leveraging these diverse datasets, we were able to thoroughly test our depth estimation and pose estimation modules across various scenarios and data types, ensuring their robustness and generalizability in endoscopic applications.

\section{Experimenental Configuration}
\label{sec:ExC}

To validate the effectiveness and generalization capabilities of our proposed MVSLAM approach, we conducted a comprehensive comparative analysis against four state-of-the-art approaches in the field of endoscopic MVSLAM and 3D reconstruction. 

EndoSLAM\footnote{\url{https://github.com/CapsuleEndoscope/EndoSLAM}} \cite{Ozyoruk2021EndoSLAMDA}, also known as Endo-SfMLearner, is an unsupervised monocular visual odometry and depth estimation approach designed specifically for endoscopic videos. 
It employs a DepthNet based on a U-Net \cite{unet} architecture with ResNet-18 encoder to predict dense disparity maps from single images. 
The PoseNet estimates six Degree of freedom relative camera poses between consecutive frames. Key innovations include a spatial attention module to focus on highly textured regions and an affine brightness transformer to handle illumination changes. 

EndoDepth\footnote{\url{https://github.com/UZ-SLAMLab/Endo-Depth-and-Motion}} \cite{Recasens2021EndoDepthandMotionRA} presents an approach for 3D reconstruction and camera motion estimation in monocular endoscopic sequences. It utilizes self-supervised depth networks, photometric tracking, and volumetric fusion to generate pseudo-RGBD frames, track camera pose, and fuse depth maps into a 3D model. 
The depth estimation uses the Monodepth2 \cite{monodepth2} network architecture with a ResNet-18 \cite{resnet} encoder. 
Camera poses are estimated by minimizing photometric error between frames.

Shao et al.\footnote{\url{https://github.com/ShuweiShao/AF-SfMLearner}}  \cite{shao2022self} propose a self-supervised framework for monocular depth and ego-motion estimation in endoscopy that introduces the concept of appearance flow to handle brightness inconsistency. Their method, also known as AFSfMlearner, comprises four modules. The Structure module (DepthNet) uses a ResNet-18 encoder-decoder architecture to estimate dense depth maps. The Motion module (PoseNet) is a lightweight network that predicts six Degree of freedom relative camera poses between frames. The Appearance module predicts appearance flow and calibrates brightness, using a structure similar to DepthNet. The Correspondence module performs automatic registration to enhance performance, with architecture similar to the appearance module. The appearance flow allows relaxation of the brightness constancy assumption, addressing severe brightness fluctuations common in endoscopic scenarios. 

OneSLAM\footnote{\url{https://github.com/arcadelab/OneSLAM}} \cite{Teufel2024} presents a generalized SLAM approach for monocular endoscopic imaging based on tracking any point. It leverages a tracking any point foundation model to obtain sparse correspondences across multiple frames. These are used to perform local bundle adjustment to jointly optimize camera poses and 3D structure. OneSLAM demonstrates strong cross-domain performance on sinus endoscopy, colonoscopy, arthroscopy and laparoscopy data without retraining.

All competitor methods were utilized as developed by their respective authors, without modifications. 
This varied baseline allows us to comprehensively assess the generalization capabilities and performance of our proposed MVSLAM approach across different endoscopic scenarios and datasets.

\subsection{Performance Metrics and Statistical Analysis}

To evaluate the performance of the SLAM approach, we employed a set of well-established metrics for depth estimation and pose estimation. For depth estimation, we selected the Absolute Relative Difference, Root Mean Square Error (RMSE), RMSE log, Squared Relative Error, and accuracy thresholds of $1.25$, $1.25^2$, and $1.25^3$, as described in~\cite{Masoumian2022MonocularDE} and presented in \tablename~\ref{table:depth_metrics}.
\begin{table}
\centering
\footnotesize
\setlength{\tabcolsep}{2pt}
\begin{tabular}{@{}p{1.8cm}p{3.5cm}p{3cm}@{}}
\toprule
\textbf{Metric} & \textbf{Formula} & \textbf{Description} \\
\midrule
Abs. Rel. Diff. & $\frac{1}{N} \sum_{i=1}^{N} \frac{|d_{i}^p - d_{i}^t|}{d_{i}^t}$ & Average of relative depth prediction errors \\
\addlinespace
Sq. Rel. & $\frac{1}{N} \sum_{i=1}^{N} \left( \frac{d_{i}^p - d_{i}^t}{d_{i}^t} \right)^2$ & Emphasizes larger errors by squaring relative error \\
\addlinespace
RMSE & $\sqrt{\frac{1}{N} \sum_{i=1}^{N} (d_{i}^p - d_{i}^t)^2}$ & Standard deviation of errors, weights larger errors more \\
\addlinespace
RMSE-Log & $\sqrt{\frac{1}{N} \sum_{i=1}^{N} (\log d_{i}^p - \log d_{i}^t)^2}$ & Evaluates errors in log scale, for wide depth ranges \\
\addlinespace
Accuracy($\delta$) & $\frac{1}{N} \sum_{i=1}^{N} \left[ \max\left(\frac{d_{i}^p}{d_{i}^t}, \frac{d_{i}^t}{d_{i}^p}\right) < \delta \right]$ & Proportion of predictions within factor $\delta$ of true values \\
\bottomrule
\end{tabular}
\caption{Key Metrics for Evaluating Depth Prediction Accuracy. $d^p$: predicted depth, $d^t$: true depth.}
\label{table:depth_metrics}
\end{table}
\begin{table}
\centering
\footnotesize
\setlength{\tabcolsep}{2pt}
\begin{tabular}{@{}p{1.5cm}p{2.9cm}p{3.8cm}@{}}
\toprule
\textbf{Metric} & \textbf{Formula} & \textbf{Description} \\
\midrule
ATE & $\| \text{trans}(E_{ij}) \|$ & Magnitude of translational pose error, capturing overall trajectory deviation \\
\addlinespace
RTE & $\| \text{trans}(E_{ij}) \|$ & Translational error over trajectory segments, assessing local consistency and drift \\
\addlinespace
RRE & $| \text{angle}(\log_{\text{SO3}}(\text{rot}(E_{ij}))) |$ & Angular deviation in rotation component, indicating rotational drift \\
\bottomrule
\end{tabular}
\caption{Comparative Overview of Trajectory Evaluation Metrics. $E_{ij}$represents the pose error between estimated and ground truth poses. The function trans($E_{ij}$) extracts the translational component, and $angle(\cdot)$ calculates the angular difference from rotational data.}
\label{table:trajectory_metrics_evo}
\end{table}

To measure the pose estimation performance, we utilized the Relative Translation Error (RTE), Relative Rotation Error (RRE), and Absolute Translation Error (ATE), as shown in \tablename~\ref{table:trajectory_metrics_evo}. 
These metrics are widely used in the literature for evaluating the accuracy of pose estimation in SLAM systems \cite{SCHMIDT2024103131}.

In addition to the performance metrics, we conducted a statistical analysis to determine the significance of the performance differences observed between the MVSLAM approach and the benchmark models across various scenarios. 
We verified that the data for each model meets the assumptions required for using an independent two-sample T-test, including normality and equal variances. 
We then performed pairwise comparisons between the MVSLAM approach and each benchmark model separately, using a significance level of $p = 0.05$.

\section{Results}
\label{sec:Res}
In this section, we present the results of our comprehensive evaluation aimed at validating the efficacy and robustness of our proposed MVSLAM approach.
Initially, in section \ref{subsec:MPEM-Res}, we provide an analysis of the monocular pose estimation module.
This is followed by section \ref{subsec:MDEM-Res}, where we explore the application and validation of the depth estimation module.
It's worth noting that our preprocessing methods for certain datasets differ from some competitors, which may lead to variations in results. For instance, we used a version of the Hamlyn dataset that had been preprocessed using libellas, as this preprocessed version has been utilized in other notable works such as \cite{Recasens2021EndoDepthandMotionRA}. 
This choice ensures consistency and comparability with existing literature in the field.
For the SCARED Dataset, we derived depth information through stereo rectification and disparity estimation, as opposed to extracting it directly from the point cloud as some other studies have done. We opted for this approach because it allows us to obtain a dense depth map, in contrast to the sparse depth information typically yielded by point cloud extraction.
It should be noted that the 3D Mapping Module (3DM) cannot be directly evaluated due to two key factors. First, the lack of ground truth 3D models for the tested environments hinders a comprehensive validation of the reconstruction quality.
Second, even if ground truth 3D models were available, the scale ambiguity inherent in the monocular pose and depth estimation modules would render the reconstructed model incomparable due to differing scales.
However, this inability to directly evaluate the 3DM does not represent a limitation of our approach. 
The 3DM employs state-of-the-art 3D reconstruction algorithms that have been rigorously tested and validated in prior work \cite{Curless1996AVM, Zhou2013DenseSR}. Furthermore, the 3D mapping and localization capabilities critically depend on the performance of the pose estimation and depth estimation modules, which are thoroughly evaluated in the following sections. The successful functioning of the overall SLAM system thus implicitly validates the efficacy of the 3DM.

\subsection{Monocular Pose Estimation Module (MPEM)}
\label{subsec:MPEM-Res}
\begin{figure*}
	\centering
	\includegraphics[width=0.8\textwidth]{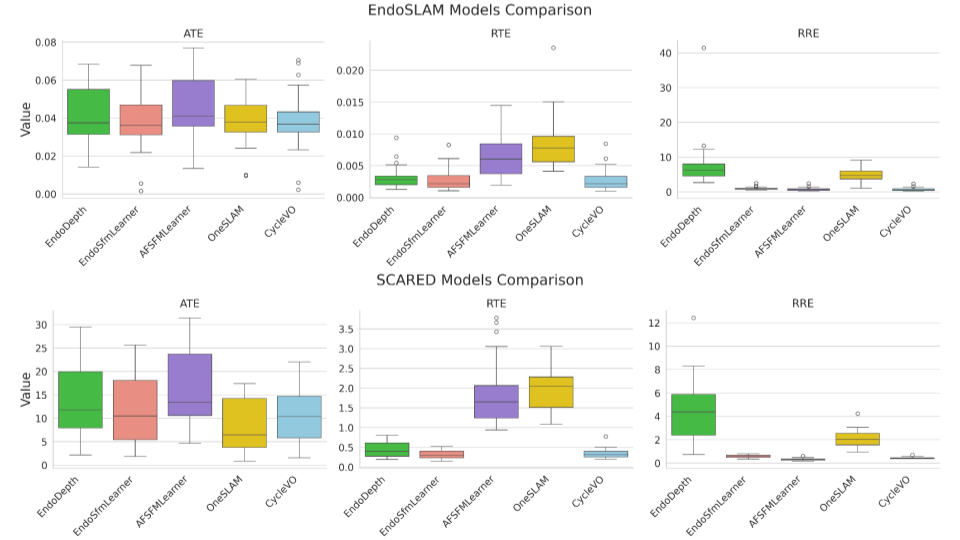}
	\caption{Comparison of the performance of EndoDepth, EndoSfmLearner, AFSfMlearner, OneSLAM and CycleVO algorithms on the SCARED and ENDOSLAM datasets. Metrics evaluated include Absolute Trajectory Error (ATE), Relative Trajectory Error (RTE), and Relative Rotation Error (RRE). Each box plot shows the distribution of the errors for the respective algorithms, highlighting the median, interquartile range, and outliers.}
	\label{fig:qualitative_results_MPEM}
\end{figure*}

\figurename~\ref{fig:qualitative_results_MPEM} presents the performances of the monocular pose estimation module measured in terms of ATE, RTE, and RRE on the two datasets (SCARED and EndoSLAM) when adopting each of the five methods, i.e., EndoSfmLearner, EndoDepth, AFSfMlearner, OneSLAM and CycleVO.
When interpreting these results, it's important to consider the training backgrounds of the methods. EndoSLAM was trained on the EndoSLAM dataset, while AFSfMLearner was trained on the SCARED dataset. This prior exposure to the test data or related datasets may introduce a potential bias in their performance, which should be taken into account when interpreting their results. In contrast, CycleVO has not been exposed to either dataset during its training.  

In the case of trajectory errors, being either an absolute or a relative error, both rows show that CycleVO and EndoSfmLearner have lower median scores and interquartile ranges than EndoDepth, AFSfMlearner and OneSLAM. 
The figure compares the three methods across the two datasets and three metrics. 
The rows represent the SCARED and ENDOSLAM datasets, while the columns show the ATE, RTE, and RRE metrics respectively.
 Regarding the statistical analysis, \tablename~\ref{tab:pose_statistical_analysis} shows the pairwise comparisons between CycleVO and the other methods. Notably, CycleVO achieves competitive performance without any prior exposure to either the EndoSLAM or SCARED datasets during its training, highlighting its strong generalization capabilities. 
It is also interesting to note that our CycleVO approach not only performs as the others but it also has the lowest inference time (\tablename~\ref{tab:inference_time}).
\begin{table}
\centering
\small
\setlength{\tabcolsep}{2pt}
\begin{tabular}{@{}cccrrr@{}}
\toprule
\textbf{} & \textbf{Metric} & \textbf{Comparison} & \textbf{T-Stat} & \textbf{P-Value} \\
\midrule
\multirow{12}{*}{\rotatebox[origin=c]{90}{SCARED}} 
& \multirow{4}{*}{ATE} & BodySLAM vs EndoDepth & -1.12 & 0.2693 \\
& & BodySLAM vs EndoSfMLearner & -0.59 & 0.5593 \\
& & BodySLAM vs OneSLAM & 1.18 & 0.2439 \\
& & BodySLAM vs AFSfMlearner & -2.44 & \textbf{0.0205} \\
\cmidrule{2-5}
& \multirow{4}{*}{RTE} & BodySLAM vs EndoDepth & -1.65 & 0.1075 \\
& & BodySLAM vs EndoSfMLearner & 0.63 & 0.5335 \\
& & BodySLAM vs OneSLAM & -11.79 & \textbf{1.81e-10} \\
& & BodySLAM vs AFSfMlearner & -7.01 & \textbf{1.20e-06} \\
\cmidrule{2-5}
& \multirow{4}{*}{RRE} & BodySLAM vs EndoDepth & -6.23 & \textbf{6.99e-06} \\
& & BodySLAM vs EndoSfMLearner & -3.53 & \textbf{0.0014} \\
& & BodySLAM vs OneSLAM & -9.00 & \textbf{3.70e-08} \\
& & BodySLAM vs AFSfMlearner & 4.17 & \textbf{0.0002} \\
\midrule
\multirow{12}{*}{\rotatebox[origin=c]{90}{ENDOSLAM}} 
& \multirow{4}{*}{ATE} & BodySLAM vs EndoDepth & -1.02 & 0.3122 \\
& & BodySLAM vs EndoSfMLearner & -0.11 & 0.9087 \\
& & BodySLAM vs OneSLAM & -0.34 & 0.7343 \\
& & BodySLAM vs AFSfMlearner & -2.68 & \textbf{0.0090} \\
\cmidrule{2-5}
& \multirow{4}{*}{RTE} & BodySLAM vs EndoDepth & -1.29 & 0.2012 \\
& & BodySLAM vs EndoSfMLearner & -0.01 & 0.9942 \\
& & BodySLAM vs OneSLAM & -9.79 & \textbf{7.24e-14} \\
& & BodySLAM vs AFSfMlearner & -7.60 & \textbf{1.73e-10} \\
\cmidrule{2-5}
& \multirow{4}{*}{RRE} & BodySLAM vs EndoDepth & -7.59 & \textbf{1.73e-09} \\
& & BodySLAM vs EndoSfMLearner & -3.85 & \textbf{0.0002} \\
& & BodySLAM vs OneSLAM & -15.39 & \textbf{5.06e-20} \\
& & BodySLAM vs AFSfMlearner & -0.63 & 0.5299 \\
\bottomrule
\end{tabular}
\caption{Comparative analysis of BodySLAM against EndoDepth, EndoSfMLearner, OneSLAM, and AFSfMlearner across different metrics (ATE, RTE, RRE) for SCARED and ENDOSLAM datasets. T-Statistic and P-Value are reported for each comparison. Bold P-Values indicate statistically significant differences (p < 0.05).}
\label{tab:pose_statistical_analysis}
\end{table}
\begin{table}
\centering
\begin{tabular}{cp{2.0cm}p{2.5cm}}  
\toprule
\textbf{Model} & \textbf{Dataset} & \textbf{ Average Time [s]} \\
\midrule
\multirow{2}{*}{EndoDepth} & EndoSLAM & 1140.56 \\
 & SCARED & 934.18 \\
\addlinespace
\multirow{2}{*}{EndoSFMLearner} & EndoSLAM & 38.95 \\
 & SCARED & 97.72 \\
\addlinespace
\multirow{2}{*}{AFSfMlearner} & EndoSLAM & 14.63 \\
 & SCARED & 59.95 \\
\addlinespace
\multirow{2}{*}{OneSLAM} & EndoSLAM & 169.77 \\
 & SCARED & 336.65 \\
\addlinespace
\multirow{2}{*}{CycleVO} & EndoSLAM & \textbf{3.32} \\
 & SCARED & \textbf{38.57} \\
\bottomrule
\end{tabular}
\vspace{0.3cm}  
\caption{Comparison of models, datasets, and execution times. This table presents the processing times for different models across two datasets, EndoSLAM and SCARED, highlighting the computational efficiency of each approach. All computations were performed on an NVIDIA RTX 3070 GPU.}
\label{tab:inference_time}
\end{table}
\begin{table}[!h]
\centering
\footnotesize
\setlength{\tabcolsep}{2pt}
\begin{tabular}{@{}lcrr@{}}
\toprule
\textbf{Metric} & \textbf{Comparison} & \textbf{T-Stat} & \textbf{P-Value} \\
\midrule
\multirow{3}{*}{\makecell[l]{Absolute Relative\\Difference}} 
& BodySLAM vs EndoDepth & -2.18 & \textbf{0.0366} \\
& BodySLAM vs EndoSfMLearner & -4.24 & \textbf{0.0002} \\
& BodySLAM vs AFSfMLearner & -12.50 & \textbf{3.55e-13} \\
\midrule
\multirow{3}{*}{RMSE} 
& BodySLAM vs EndoDepth & -1.99 & 0.0550 \\
& BodySLAM vs EndoSfMLearner & -2.86 & \textbf{0.0069} \\
& BodySLAM vs AFSfMLearner & -6.26 & \textbf{2.73e-06} \\
\midrule
\multirow{3}{*}{RMSE Log} 
& BodySLAM vs EndoDepth & -2.13 & \textbf{0.0400} \\
& BodySLAM vs EndoSfMLearner & -4.42 & \textbf{9.94e-05} \\
& BodySLAM vs AFSfMLearner & -17.08 & \textbf{1.29e-15} \\
\midrule
\multirow{3}{*}{\makecell[l]{Squared Relative\\Error}} 
& BodySLAM vs EndoDepth & -2.15 & \textbf{0.0422} \\
& BodySLAM vs EndoSfMLearner & -2.71 & \textbf{0.0105} \\
& BodySLAM vs AFSfMLearner & -4.80 & \textbf{0.0001} \\
\midrule
\multirow{3}{*}{accuracy (1.25)} 
& BodySLAM vs EndoDepth & 5.14 & \textbf{4.42e-05} \\
& BodySLAM vs EndoSfMLearner & 2.55 & \textbf{0.0151} \\
& BodySLAM vs AFSfMLearner & 14.17 & \textbf{2.55e-13} \\
\midrule
\multirow{3}{*}{accuracy (1.25)²} 
& BodySLAM vs EndoDepth & 4.12 & \textbf{0.0006} \\
& BodySLAM vs EndoSfMLearner & 1.54 & 0.1396 \\
& BodySLAM vs AFSfMLearner & 10.38 & \textbf{4.62e-09} \\
\midrule
\multirow{3}{*}{accuracy (1.25)³} 
& BodySLAM vs EndoDepth & 3.56 & \textbf{0.0022} \\
& BodySLAM vs EndoSfMLearner & 1.37 & 0.1886 \\
& BodySLAM vs AFSfMLearner & 14.10 & \textbf{3.62e-11} \\
\bottomrule
\end{tabular}
\caption{Statistical Performance Analysis of the Monocular Depth Estimation Module on the Hamlyn Dataset. The table details outcomes for Absolute Relative Difference, Root Mean Square Error (RMSE), RMSE Log, Squared Relative Error, and accuracy thresholds (accuracy < 1.25, accuracy < 1.25², accuracy < 1.25³). Results include pairwise comparisons of BodySLAM versus EndoDepth, EndoSfMLearner, and AFSfMLearner, assessed through t-tests with a significance level of p < 0.05.}
\label{tab:statistical_analysis_MDEM_Hamlyn}
\end{table}
\begin{table}[thbp]
\centering
\footnotesize
\setlength{\tabcolsep}{2pt}
\begin{tabular}{@{}lcrr@{}}
\toprule
\textbf{Metric} & \textbf{Comparison} & \textbf{T-Stat} & \textbf{P-Value} \\
\midrule
\multirow{3}{*}{\makecell[l]{Absolute Relative\\Difference}} 
& BodySLAM vs EndoDepth & -9.69 & \textbf{1.81e-09} \\
& BodySLAM vs EndoSfMLearner & -4.14 & \textbf{1.99e-04} \\
& BodySLAM vs AFSfMlearner & -4.36 & \textbf{3.39e-04} \\
\midrule
\multirow{3}{*}{RMSE} 
& BodySLAM vs EndoDepth & -1.01 & 0.3212 \\
& BodySLAM vs EndoSfMLearner & -0.25 & 0.8007 \\
& BodySLAM vs AFSfMlearner & -0.78 & 0.4392 \\
\midrule
\multirow{3}{*}{RMSE Log} 
& BodySLAM vs EndoDepth & -14.99 & \textbf{2.72e-13} \\
& BodySLAM vs EndoSfMLearner & -8.35 & \textbf{1.05e-09} \\
& BodySLAM vs AFSfMlearner & -13.13 & \textbf{4.65e-11} \\
\midrule
\multirow{3}{*}{\makecell[l]{Squared Relative\\Error}} 
& BodySLAM vs EndoDepth & -0.85 & 0.4069 \\
& BodySLAM vs EndoSfMLearner & 0.31 & 0.7601 \\
& BodySLAM vs AFSfMlearner & -0.35 & 0.7305 \\
\midrule
\multirow{3}{*}{accuracy (1.25)} 
& BodySLAM vs EndoDepth & 13.10 & \textbf{3.61e-11} \\
& BodySLAM vs EndoSfMLearner & 5.82 & \textbf{6.53e-06} \\
& BodySLAM vs AFSfMlearner & 9.36 & \textbf{1.69e-08} \\
\midrule
\multirow{3}{*}{accuracy (1.25)²} 
& BodySLAM vs EndoDepth & 10.13 & \textbf{7.29e-09} \\
& BodySLAM vs EndoSfMLearner & 2.56 & \textbf{0.0195} \\
& BodySLAM vs AFSfMlearner & 6.26 & \textbf{6.59e-06} \\
\midrule
\multirow{3}{*}{accuracy (1.25)³} 
& BodySLAM vs EndoDepth & 8.53 & \textbf{9.75e-08} \\
& BodySLAM vs EndoSfMLearner & 1.26 & 0.2228 \\
& BodySLAM vs AFSfMlearner & 5.47 & \textbf{3.40e-05} \\
\bottomrule
\end{tabular}
\caption{Statistical Performance Analysis of the Monocular Depth Estimation Module on the SCARED Dataset. The table details outcomes for Absolute Relative Difference, Root Mean Square Error (RMSE), RMSE Log, Squared Relative Error, and accuracy thresholds (accuracy < 1.25, accuracy < 1.25², accuracy < 1.25³). Results include pairwise comparisons of BodySLAM versus EndoDepth, EndoSfMLearner, and AFSfMlearner, assessed through t-tests with a significance level of p < 0.05.}
\label{tab:statistical_analysis_MDEM_SCARED}
\end{table}

\begin{figure*}
	\centering
	\includegraphics[width=0.8\textwidth]{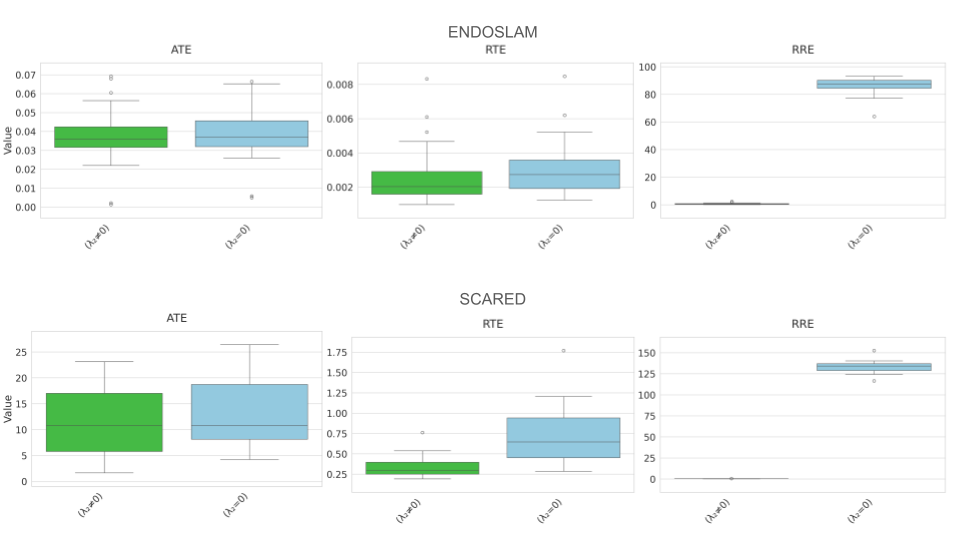}
	\caption{Ablation study of cycle loss term modifications. This figure presents the impact of different cycle loss term settings ( $\lambda_2 \neq 0$ vs $\lambda_2 = 0$) on the performance of models evaluated on the SCARED and ENDOSLAM datasets. The metrics assessed include Absolute Trajectory Error (ATE), Relative Trajectory Error (RTE), and Relative Rotation Error (RRE). Each box plot illustrates the distribution of the errors for the respective models, highlighting the median, interquartile range, and outliers.}
	\label{fig:ablation_study_pose}
\end{figure*}

Turning our attention to RRE, i.e., a score measuring the angular deviation in the rotation component, we notice that CycleVO outperforms both EndoSfmLearner, EndoDepth, AFSfMlearner and OneSLAM on both datasets (both rows and last column of \figurename~\ref{fig:qualitative_results_MPEM}).
Furthermore, these performance differences are also statistically significant (last rows for both datasets of \figurename~\ref{fig:qualitative_results_MPEM}).
To deepen the results we also perform an ablation study that aims to investigate the effectiveness of the new terms that we introduced for the cycle loss (section~\ref{subsec:MPEM}). 
To this end, we ran four experiments, two per dataset, setting $\lambda_2 \neq 0$, $\lambda_2 = 0$ (\figurename~\ref{fig:ablation_study_pose}). 
Both in the case of trajectory errors, being either an absolute or a relative error, and the relative rotation error, both rows show that the model with the new term ($\lambda_2 \neq 0$) has lower median scores and interquartile ranges than the counterpart without the custom term ($\lambda_2 = 0$).

\subsection{Monocular Depth Estimation Module (MDEM)}
\label{subsec:MDEM-Res}
\begin{figure*}
	\centering
	\makebox[\textwidth]{\includegraphics[width=0.8\textwidth]{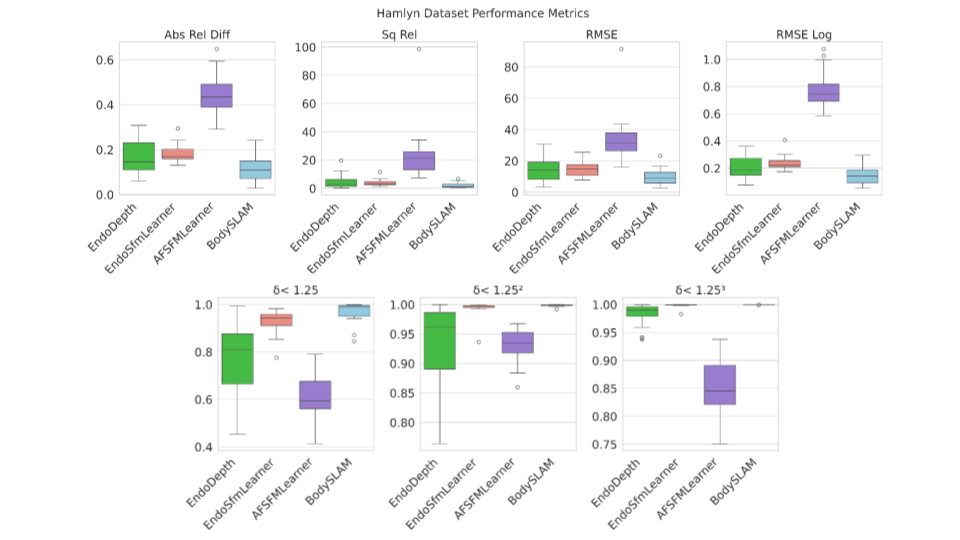}}
        \vspace{-20pt} 
	\caption{Comparison of depth estimation performance for EndoDepth, EndoSfmLearner, AFSfMlearner and BodySLAM across various metrics on Hamlyn Dataset. Box plots show that BodySLAM achieves the best performance with the lowest median values for absolute relative difference, RMSE, RMSE log, and squared relative difference, and the highest median values for accuracy thresholds $\delta < 1.25$, $\delta < 1.25^2$, and $\delta < 1.25^3$. EndoSfmLearner demonstrates the second best performance, while EndoDepth shows the poorest results among the three methods compared.}
	\label{fig:qualitative_results_MDEM_Hamlyn}
\end{figure*}

\begin{figure*}
	\centering
	\makebox[\textwidth]{\includegraphics[width=0.8\textwidth]{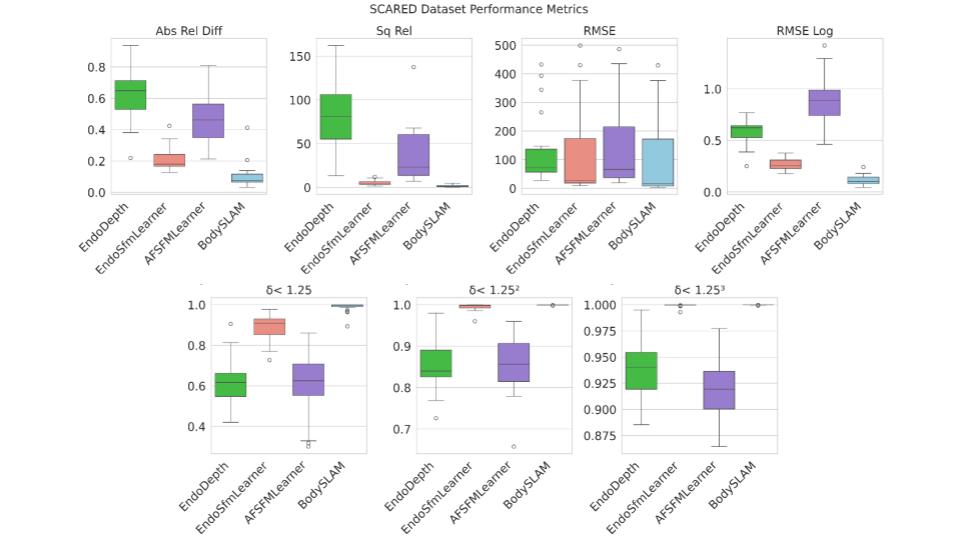}}
        \vspace{-20pt} 
	\caption{Comparison of depth estimation performance for EndoDepth, EndoSfmLearner, AFSfMlearner and BodySLAM across various metrics on Hamlyn Dataset. Box plots show that BodySLAM achieves the best performance with the lowest median values for absolute relative difference, RMSE log, and squared relative difference, and the highest median values for accuracy thresholds $\delta < 1.25$, $\delta < 1.25^2$, and $\delta < 1.25^3$. EndoSfmLearner demonstrates the second best performance, while EndoDepth shows the poorest results among the three methods compared. Outliers beyond 1.5 times the interquartile range from the quartiles have been excluded to enhance clarity and focus on the central trends of the data.}
	\label{fig:qualitative_results_MDEM_SCARED}
\end{figure*}

\figurename~\ref{fig:qualitative_results_MDEM_Hamlyn}, \ref{fig:qualitative_results_MDEM_SCARED} present the performances of the monocular depth estimation module, BodySLAM, compared to EndoSfmLearner, EndoDepth and AFSfMLearner respectively on the Hamlyn and SCARED datasets. 
The figures compare the three methods across multiple evaluation metrics for each dataset. The x-axis represents the different methods (EndoSfmLearner, EndoDepth, AFSfMLearner and BodySLAM), while the y-axis shows the values for the evaluation metrics.

It's worth noting that OneSLAM is not included in this comparison. This is because OneSLAM produces sparse depth maps, while the other methods generate dense depth maps. 
The evaluation metrics used in this study are designed for dense depth maps, making a direct comparison with OneSLAM's sparse output impossible.
When interpreting these results, it's important to consider the training backgrounds of the methods. 
EndoDepth was trained on the Hamlyn dataset, while AFSfMLearner was trained on the SCARED dataset. 
This prior exposure to the test data or related datasets may introduce a potential bias in their performance, which should be taken into account when interpreting their results. 
In contrast, BodySLAM has not been exposed to either dataset during its training, which highlights its generalization capabilities.

In the case of the Hamlyn dataset (\figurename~\ref{fig:qualitative_results_MDEM_Hamlyn}), BodySLAM outperforms all competitors across all metrics. 
For Absolute relative difference, RMSE, RMSE log and Squared Relative error (first row, \figurename~\ref{fig:qualitative_results_MDEM_Hamlyn}), BodySLAM exhibits lower median values and smaller interquartile ranges compared to the other methods. Regarding the threshold accuracy (second row, \figurename~\ref{fig:qualitative_results_MDEM_Hamlyn}), where values close to or equal to 1 are considered optimal, BodySLAM shows a median closest to this optimal value and a smaller interquartile range compared to the others.
Similarly, for the SCARED dataset (\figurename~\ref{fig:qualitative_results_MDEM_SCARED}), BodySLAM outperforms all competitors across all metrics. As with the Hamlyn dataset, BodySLAM demonstrates lower median values and smaller interquartile ranges for Absolute relative difference, RMSE, RMSE log and Squared Relative error (first row, \figurename~\ref{fig:qualitative_results_MDEM_SCARED}). 
For the threshold accuracy (second row, \figurename~\ref{fig:qualitative_results_MDEM_SCARED}), BodySLAM median is closest to the optimal value of 1, with a smaller interquartile range compared to the other methods.
To deepen the results we also performed a statistical analysis presented in \tablename~\ref{tab:statistical_analysis_MDEM_Hamlyn} for the Hamlyn dataset and in \tablename~\ref{tab:statistical_analysis_MDEM_SCARED} for the SCARED Dataset. 
The statistical analysis presented on the Hamlyn dataset shows that BodySLAM outperforms the other depth estimation models, except for the threshold accuracy $(1.25)^{2}$ and $(1.25)^{3}$, where BodySLAM does not significantly outperform EndoSfmLearner.
For the SCARED dataset, BodySLAM outperforms the other depth estimation models in most metrics. However, there are a few exceptions: in RMSE, BodySLAM does not statistically outperform EndoSfmLearner; in Squared Relative Error, BodySLAM does not statistically outperform both EndoDepth and EndoSfmLearner; and in Accuracy $(1.25)^{3}$, BodySLAM does not statistically outperform EndoSfmLearner.

\section{Conclusions}
\label{sec:Disc&ftw}
This study presented a robust deep learning-based SLAM approach that can effectively operate across various endoscopic surgical settings, including laparoscopy, gastroscopy, and colonoscopy. 
We addressed the challenges posed by hardware limitations and environmental variations by integrating deep learning models with strong generalization capabilities into the framework.
Our results demonstrate that our novel unsupervised pose estimation method, CycleVO, exhibits robust generalization capabilities and competitive performance compared to existing state-of-the-art pose estimation techniques, while maintaining the lowest inference time. This makes CycleVO a promising solution for real-time surgical applications. An ablation study conducted on the pose estimation module highlighted the effectiveness of new terms introduced for cycle loss, with the model incorporating the custom term consistently outperforming its counterpart without the term. These findings were confirmed by statistical analysis revealing that the CycleVO approach performed comparably to other methods while having the lowest inference time.
Furthermore, we demonstrated that Zoe's exceptional generalization abilities, previously showcased in various imaging contexts, can be effectively adapted to the endoscopic field. When compared to current state-of-the-art depth estimation algorithms in endoscopy, Zoe exhibited superior performance, confirming its potential for SLAM applications in surgical environments. These findings are consistent with the results reported by \cite{Han2024DepthAI}. Statistical analysis revealed that Zoe significantly outperformed other depth estimation models in most metrics, with only a few exceptions where the differences were not statistically significant. \\
Despite these promising results, our study has four limitations. 
First, although we validated our approach on three relevant endoscopic scenarios, its performance in other settings remains to be investigated. Future research should focus on collecting diverse datasets from a wider range of endoscopic procedures and fine-tuning the deep learning models to adapt to the unique characteristics of each domain, which could help extend the framework's applicability.
Second, the impact of varying illumination conditions and the presence of fluids or debris on the approach's performance needs further examination. Investigating domain adaptation techniques and developing specialized modules for handling procedure-specific challenges, such as varying illumination and the presence of fluids, could improve the approach's robustness.
Third, the scale ambiguity problem inherent to monocular cameras may affect both the depth estimation and pose estimation modules, resulting in a 3D model that has a scale that differs from the real scale. Exploring methods to mitigate the scale ambiguity problem, such as incorporating additional sensors or prior knowledge, could enhance the accuracy of depth estimation and pose estimation, and enable more reliable 3D reconstruction validation.  
Fourth, we have in-silico validated our approach and we are aware that the practical application of the approach in real-world surgical scenarios requires extensive validation and collaboration with medical experts. Conducting rigorous clinical trials involving diverse patient populations and surgical conditions, in close collaboration with medical experts, will be essential for validating the approach's performance and ensuring its seamless integration into the surgical workflow.
In spite of these limitations, the proposed SLAM approach has the potential to significantly improve the accuracy and efficiency of endoscopic procedures by providing surgeons with enhanced depth perception and 3D reconstruction capabilities. By addressing the challenges posed by hardware limitations and environmental variations, our deep learning-based framework represents a promising step towards overcoming the limitations of current endoscopic surgical practices. Future research efforts, guided by the identified limitations and potential improvements, can further refine and validate this approach, ultimately leading to its successful integration into clinical practice and improved surgical outcomes for patients.

\printcredits
\section*{Declaration of competing interest}
The authors declare that they have no known competing financial interests or personal relationships that could have appeared to
influence the work reported in this paper.

\section*{Acknowledgements}
Guido Manni is a Ph.D. student enrolled in the National
Ph.D. in Artificial Intelligence, XXXVII cycle, course on Health and life
sciences, organized by Università Campus Bio-Medico di Roma. \\
This work was partially founded by: i) Piano Nazionale Ripresa e Resilienza (PNRR) - HEAL ITALIA Extended Partnership - SPOKE 2 Cascade Call - "Intelligent Health" with the project BISTOURY - 3D-guided roBotIc Surgery based on advanced navigaTiOn systems and aUgmented viRtual realitY (CUP: J33C22002920006); ii) UCBM University Strategic Projects 2023 with the Proof of Concept (PoC) project BONE - Cooperative Robotic System for spinal surgery; iii) PNRR MUR project PE0000013-FAIR.
Resources are provided by the National Academic Infrastructure for Supercomputing in Sweden (NAISS) and the Swedish National Infrastructure for Computing (SNIC) at Alvis @ C3S.

\bibliographystyle{elsarticle-num}

\bibliography{references}

\begin{thebibliography}{10}
\expandafter\ifx\csname url\endcsname\relax
  \def\url#1{\texttt{#1}}\fi
\expandafter\ifx\csname urlprefix\endcsname\relax\def\urlprefix{URL }\fi
\expandafter\ifx\csname href\endcsname\relax
  \def\href#1#2{#2} \def\path#1{#1}\fi

\bibitem{SCHMIDT2024103131}
A.~Schmidt, O.~Mohareri, S.~DiMaio, M.~C. Yip, S.~E. Salcudean,
  \href{https://www.sciencedirect.com/science/article/pii/S1361841524000562}{Tracking
  and mapping in medical computer vision: A review}, Medical Image Analysis 94
  (2024) 103131.
\newblock \href {https://doi.org/https://doi.org/10.1016/j.media.2024.103131}
  {\path{doi:https://doi.org/10.1016/j.media.2024.103131}}.
\newline\urlprefix\url{https://www.sciencedirect.com/science/article/pii/S1361841524000562}

\bibitem{DURRANI1995237}
A.~F. Durrani, G.~M. Preminger,
  \href{https://www.sciencedirect.com/science/article/pii/001048259500001K}{Three-dimensional
  video imaging for endoscopic surgery}, Computers in Biology and Medicine
  25~(2) (1995) 237--247, virtual Reality for Medicine.
\newblock \href {https://doi.org/https://doi.org/10.1016/0010-4825(95)00001-K}
  {\path{doi:https://doi.org/10.1016/0010-4825(95)00001-K}}.
\newline\urlprefix\url{https://www.sciencedirect.com/science/article/pii/001048259500001K}

\bibitem{Stewart2007ThePO}
L.~Stewart, L.~W. Way, \href{https://doi.org/10.1177/154193120705101103}{The
  prevention of laparoscopic bile duct injuries: An analysis of 300 cases of
  from a human factors and cognitive psychology perspective}, Proceedings of
  the Human Factors and Ergonomics Society Annual Meeting 51~(11) (2007)
  617--620.
\newblock \href
  {http://arxiv.org/abs/https://doi.org/10.1177/154193120705101103}
  {\path{arXiv:https://doi.org/10.1177/154193120705101103}}, \href
  {https://doi.org/10.1177/154193120705101103}
  {\path{doi:10.1177/154193120705101103}}.
\newline\urlprefix\url{https://doi.org/10.1177/154193120705101103}

\bibitem{Smith2014EffectOP}
R.~Smith, K.~Schwab, A.~Day, T.~Rockall, K.~Ballard, M.~Bailey, I.~Jourdan,
  \href{https://doi.org/10.1002/bjs.9601}{{Effect of passive polarizing
  three-dimensional displays on surgical performance for experienced
  laparoscopic surgeons}}, British Journal of Surgery 101~(11) (2014)
  1453--1459.
\newblock \href
  {http://arxiv.org/abs/https://academic.oup.com/bjs/article-pdf/101/11/1453/36619181/bjs9601.pdf}
  {\path{arXiv:https://academic.oup.com/bjs/article-pdf/101/11/1453/36619181/bjs9601.pdf}},
  \href {https://doi.org/10.1002/bjs.9601} {\path{doi:10.1002/bjs.9601}}.
\newline\urlprefix\url{https://doi.org/10.1002/bjs.9601}

\bibitem{Vettoretto2018WhyLM}
N.~Vettoretto, E.~Foglia, L.~Ferrario, A.~Arezzo, R.~Cirocchi, G.~Cocorullo,
  G.~Currò, D.~Marchi, G.~Portale, C.~Gerardi, U.~Nocco, M.~Tringali,
  G.~Anania, M.~Piccoli, G.~Silecchia, M.~Morino, A.~Valeri, E.~Lettieri,
  \href{https://doi.org/10.1007/s00464-017-6006-y}{Why laparoscopists may opt
  for three-dimensional view: a summary of the full hta report on 3d versus 2d
  laparoscopy by s.i.c.e. (società italiana di chirurgia endoscopica e nuove
  tecnologie)}, Surgical Endoscopy 32~(6) (2018) 2986--2993.
\newblock \href {https://doi.org/10.1007/s00464-017-6006-y}
  {\path{doi:10.1007/s00464-017-6006-y}}.
\newline\urlprefix\url{https://doi.org/10.1007/s00464-017-6006-y}

\bibitem{Srensen2015ThreedimensionalVT}
S.~M.~D. Sørensen, M.~M. Savran, L.~Konge, F.~Bjerrum,
  \href{https://doi.org/10.1007/s00464-015-4189-7}{Three-dimensional versus
  two-dimensional vision in laparoscopy: a systematic review}, Surgical
  Endoscopy 30~(1) (2016) 11--23.
\newblock \href {https://doi.org/10.1007/s00464-015-4189-7}
  {\path{doi:10.1007/s00464-015-4189-7}}.
\newline\urlprefix\url{https://doi.org/10.1007/s00464-015-4189-7}

\bibitem{Davies2020ThreedimensionalVT}
S.~Davies, M.~Ghallab, S.~Hajibandeh, S.~Hajibandeh, S.~Addison,
  \href{https://doi.org/10.1007/s00423-020-01909-9}{Three-dimensional versus
  two-dimensional imaging during laparoscopic cholecystectomy: a systematic
  review and meta-analysis of randomised controlled trials}, Langenbeck's
  Archives of Surgery 405~(5) (2020) 563--572.
\newblock \href {https://doi.org/10.1007/s00423-020-01909-9}
  {\path{doi:10.1007/s00423-020-01909-9}}.
\newline\urlprefix\url{https://doi.org/10.1007/s00423-020-01909-9}

\bibitem{Schwab2017EvolutionOS}
K.~Schwab, R.~Smith, V.~Brown, M.~Whyte, I.~Jourdan,
  \href{https://doi.org/10.4253/wjge.v9.i8.368}{Evolution of stereoscopic
  imaging in surgery and recent advances}, World Journal of Gastrointestinal
  Endoscopy 9~(8) (2017) 368--377.
\newblock \href {https://doi.org/10.4253/wjge.v9.i8.368}
  {\path{doi:10.4253/wjge.v9.i8.368}}.
\newline\urlprefix\url{https://doi.org/10.4253/wjge.v9.i8.368}

\bibitem{Bergen2016StitchingAS}
T.~Bergen, T.~Wittenberg,
  \href{https://doi.org/10.1109/JBHI.2014.2384134}{Stitching and surface
  reconstruction from endoscopic image sequences: A review of applications and
  methods}, IEEE Journal of Biomedical and Health Informatics 20~(1) (2016)
  304--321.
\newblock \href {https://doi.org/10.1109/JBHI.2014.2384134}
  {\path{doi:10.1109/JBHI.2014.2384134}}.
\newline\urlprefix\url{https://doi.org/10.1109/JBHI.2014.2384134}

\bibitem{Bano2022PlacentalVH}
S.~Bano, F.~Vasconcelos, A.~L. David, J.~Deprest, D.~Stoyanov,
  \href{https://doi.org/10.1080/21681163.2022.2154278}{Placental vessel-guided
  hybrid framework for fetoscopic mosaicking}, Computer Methods in Biomechanics
  and Biomedical Engineering: Imaging \& Visualization 11~(4) (2023)
  1166--1171.
\newblock \href
  {http://arxiv.org/abs/https://doi.org/10.1080/21681163.2022.2154278}
  {\path{arXiv:https://doi.org/10.1080/21681163.2022.2154278}}, \href
  {https://doi.org/10.1080/21681163.2022.2154278}
  {\path{doi:10.1080/21681163.2022.2154278}}.
\newline\urlprefix\url{https://doi.org/10.1080/21681163.2022.2154278}

\bibitem{Li2021GloballyOF}
L.~Li, S.~Bano, J.~Deprest, A.~L. David, D.~Stoyanov, F.~Vasconcelos,
  \href{https://doi.org/10.1109/LRA.2021.3100938}{Globally optimal fetoscopic
  mosaicking based on pose graph optimisation with affine constraints}, IEEE
  Robotics and Automation Letters 6~(4) (2021) 7831--7838.
\newblock \href {https://doi.org/10.1109/LRA.2021.3100938}
  {\path{doi:10.1109/LRA.2021.3100938}}.
\newline\urlprefix\url{https://doi.org/10.1109/LRA.2021.3100938}

\bibitem{Cheema2019ImageAlignedDL}
M.~N. Cheema, A.~Nazir, B.~Sheng, P.~Li, J.~Qin, J.~Kim, D.~D. Feng,
  \href{https://doi.org/10.1109/TBME.2018.2884319}{Image-aligned dynamic liver
  reconstruction using intra-operative field of views for minimal invasive
  surgery}, IEEE Transactions on Biomedical Engineering 66~(8) (2019)
  2163--2173.
\newblock \href {https://doi.org/10.1109/TBME.2018.2884319}
  {\path{doi:10.1109/TBME.2018.2884319}}.
\newline\urlprefix\url{https://doi.org/10.1109/TBME.2018.2884319}

\bibitem{Zhang20213DRO}
S.~Zhang, L.~Zhao, S.~Huang, R.~Ma, B.~Hu, Q.~Hao,
  \href{https://doi.org/10.1109/ICRA48506.2021.9561772}{3d reconstruction of
  deformable colon structures based on preoperative model and deep neural
  network}, in: 2021 IEEE International Conference on Robotics and Automation
  (ICRA), 2021, pp. 1875--1881.
\newblock \href {https://doi.org/10.1109/ICRA48506.2021.9561772}
  {\path{doi:10.1109/ICRA48506.2021.9561772}}.
\newline\urlprefix\url{https://doi.org/10.1109/ICRA48506.2021.9561772}

\bibitem{Widya2019WholeS3}
A.~R. Widya, Y.~Monno, M.~Okutomi, S.~Suzuki, T.~Gotoda, K.~Miki,
  \href{https://doi.org/10.1109/JTEHM.2019.2946802}{Whole stomach 3d
  reconstruction and frame localization from monocular endoscope video}, IEEE
  Journal of Translational Engineering in Health and Medicine 7 (2019) 1--10.
\newblock \href {https://doi.org/10.1109/JTEHM.2019.2946802}
  {\path{doi:10.1109/JTEHM.2019.2946802}}.
\newline\urlprefix\url{https://doi.org/10.1109/JTEHM.2019.2946802}

\bibitem{Makki2023EllipticalSD}
K.~Makki, K.~Chandelon, A.~Bartoli,
  \href{https://doi.org/10.1007/s11548-023-02904-3}{Elliptical specularity
  detection in endoscopy with application to normal reconstruction},
  International Journal of Computer Assisted Radiology and Surgery 18~(7)
  (2023) 1323--1328.
\newblock \href {https://doi.org/10.1007/s11548-023-02904-3}
  {\path{doi:10.1007/s11548-023-02904-3}}.
\newline\urlprefix\url{https://doi.org/10.1007/s11548-023-02904-3}

\bibitem{Zhang2022SLAMTKARI}
S.~Zhang, L.~Zhao, S.~Huang, H.~Wang, Q.~Luo, Q.~Hao,
  \href{https://doi.org/10.1007/978-3-031-16449-1_13}{Slam-tka: Real-time
  intra-operative measurement of tibial resection plane in conventional total
  knee arthroplasty}, in: L.~Wang, Q.~Dou, P.~T. Fletcher, S.~Speidel, S.~Li
  (Eds.), Medical Image Computing and Computer Assisted Intervention -- MICCAI
  2022, Springer Nature Switzerland, Cham, 2022, pp. 126--135.
\newblock \href {https://doi.org/10.1007/978-3-031-16449-1_13}
  {\path{doi:10.1007/978-3-031-16449-1_13}}.
\newline\urlprefix\url{https://doi.org/10.1007/978-3-031-16449-1_13}

\bibitem{ORBSLAM3_TRO}
C.~Campos, R.~Elvira, J.~J. Gomez, J.~M.~M. Montiel, J.~D. Tard\'os,
  {ORB-SLAM3}: An accurate open-source library for visual, visual-inertial and
  multi-map {SLAM}, IEEE Transactions on Robotics 37~(6) (2021) 1874--1890.

\bibitem{Longuet-Higgins1981}
H.~C. Longuet-Higgins, \href{https://doi.org/10.1038/293133a0}{A computer
  algorithm for reconstructing a scene from two projections}, Nature 293~(5828)
  (1981) 133--135.
\newblock \href {https://doi.org/10.1038/293133a0}
  {\path{doi:10.1038/293133a0}}.
\newline\urlprefix\url{https://doi.org/10.1038/293133a0}

\bibitem{Schle2022AMS}
J.~Schüle, J.~Haag, P.~Somers, C.~Veil, C.~Tarín, O.~Sawodny,
  \href{https://doi.org/10.1109/AIM52237.2022.9863308}{A model-based
  simultaneous localization and mapping approach for deformable bodies}, in:
  2022 IEEE/ASME International Conference on Advanced Intelligent Mechatronics
  (AIM), 2022, pp. 607--612.
\newblock \href {https://doi.org/10.1109/AIM52237.2022.9863308}
  {\path{doi:10.1109/AIM52237.2022.9863308}}.
\newline\urlprefix\url{https://doi.org/10.1109/AIM52237.2022.9863308}

\bibitem{Li2019SuPerAS}
Y.~Li, F.~Richter, J.~Lu, E.~K. Funk, R.~K. Orosco, J.~Zhu, M.~C. Yip,
  \href{https://doi.org/10.1109/LRA.2020.2970659}{Super: A surgical perception
  framework for endoscopic tissue manipulation with surgical robotics}, IEEE
  Robotics and Automation Letters 5~(2) (2020) 2294--2301.
\newblock \href {https://doi.org/10.1109/LRA.2020.2970659}
  {\path{doi:10.1109/LRA.2020.2970659}}.
\newline\urlprefix\url{https://doi.org/10.1109/LRA.2020.2970659}

\bibitem{Liu2023SurfaceDT}
Z.~Liu, W.~Gao, J.~Zhu, Z.~Yu, Y.~Fu,
  \href{https://www.sciencedirect.com/science/article/pii/S1361841523000361}{Surface
  deformation tracking in monocular laparoscopic video}, Medical Image Analysis
  86 (2023) 102775.
\newblock \href {https://doi.org/https://doi.org/10.1016/j.media.2023.102775}
  {\path{doi:https://doi.org/10.1016/j.media.2023.102775}}.
\newline\urlprefix\url{https://www.sciencedirect.com/science/article/pii/S1361841523000361}

\bibitem{Recasens2021EndoDepthandMotionRA}
D.~Recasens, J.~Lamarca, J.~M. Fácil, J.~M.~M. Montiel, J.~Civera,
  \href{https://doi.org/10.1109/LRA.2021.3095528}{Endo-depth-and-motion:
  Reconstruction and tracking in endoscopic videos using depth networks and
  photometric constraints}, IEEE Robotics and Automation Letters 6~(4) (2021)
  7225--7232.
\newblock \href {https://doi.org/10.1109/LRA.2021.3095528}
  {\path{doi:10.1109/LRA.2021.3095528}}.
\newline\urlprefix\url{https://doi.org/10.1109/LRA.2021.3095528}

\bibitem{Ozyoruk2021EndoSLAMDA}
K.~B. Ozyoruk, G.~I. Gokceler, T.~L. Bobrow, G.~Coskun, K.~Incetan,
  Y.~Almalioglu, F.~Mahmood, E.~Curto, L.~Perdigoto, M.~Oliveira, H.~Sahin,
  H.~Araujo, H.~Alexandrino, N.~J. Durr, H.~B. Gilbert, M.~Turan,
  \href{https://www.sciencedirect.com/science/article/pii/S1361841521001043}{Endoslam
  dataset and an unsupervised monocular visual odometry and depth estimation
  approach for endoscopic videos}, Medical Image Analysis 71 (2021) 102058.
\newblock \href {https://doi.org/https://doi.org/10.1016/j.media.2021.102058}
  {\path{doi:https://doi.org/10.1016/j.media.2021.102058}}.
\newline\urlprefix\url{https://www.sciencedirect.com/science/article/pii/S1361841521001043}

\bibitem{shao2022self}
Shao, Shuwei, Pei, Zhongcai, Chen, Weihai, Zhu, Wentao, Wu, Xingming, Sun,
  Dianmin, B.~Zhang, Self-supervised monocular depth and ego-motion estimation
  in endoscopy: Appearance flow to the rescue, Medical image analysis 77 (2022)
  102338.

\bibitem{Teufel2024}
T.~Teufel, H.~Shu, R.~D. Soberanis-Mukul, J.~E. Mangulabnan, M.~Sahu, S.~S.
  Vedula, M.~Ishii, G.~Hager, R.~H. Taylor, M.~Unberath,
  \href{https://doi.org/10.1007/s11548-024-03171-6}{Oneslam to map them all: a
  generalized approach to slam for monocular endoscopic imaging based on
  tracking any point}, International Journal of Computer Assisted Radiology and
  Surgery 19~(7) (2024) 1259--1266.
\newblock \href {https://doi.org/10.1007/s11548-024-03171-6}
  {\path{doi:10.1007/s11548-024-03171-6}}.
\newline\urlprefix\url{https://doi.org/10.1007/s11548-024-03171-6}

\bibitem{zoe}
S.~Bhat, R.~Birkl, D.~Wofk, P.~Wonka, M.~Mueller, Zoedepth: Zero-shot transfer
  by combining relative and metric depth (02 2023).

\bibitem{EndoVis2019}
Stereo correspondence and reconstruction of endoscopic data sub-challenge,
  \url{https://endovissub2019-scared.grand-challenge.org/}, accessed:
  2024-06-17 (2019).

\bibitem{Rublee2011ORBAE}
E.~Rublee, V.~Rabaud, K.~Konolige, G.~Bradski,
  \href{https://doi.org/10.1109/ICCV.2011.6126544}{Orb: An efficient
  alternative to sift or surf}, in: 2011 International Conference on Computer
  Vision, 2011, pp. 2564--2571.
\newblock \href {https://doi.org/10.1109/ICCV.2011.6126544}
  {\path{doi:10.1109/ICCV.2011.6126544}}.
\newline\urlprefix\url{https://doi.org/10.1109/ICCV.2011.6126544}

\bibitem{Lowe2004DistinctiveIF}
D.~G. Lowe,
  \href{https://doi.org/10.1023/B:VISI.0000029664.99615.94}{Distinctive image
  features from scale-invariant keypoints}, International Journal of Computer
  Vision 60~(2) (2004) 91--110.
\newblock \href {https://doi.org/10.1023/B:VISI.0000029664.99615.94}
  {\path{doi:10.1023/B:VISI.0000029664.99615.94}}.
\newline\urlprefix\url{https://doi.org/10.1023/B:VISI.0000029664.99615.94}

\bibitem{Bay2006SURFSU}
H.~Bay, T.~Tuytelaars, L.~Van~Gool,
  \href{https://doi.org/10.1007/11744023_32}{Surf: Speeded up robust features},
  in: A.~Leonardis, H.~Bischof, A.~Pinz (Eds.), Computer Vision -- ECCV 2006,
  Springer Berlin Heidelberg, Berlin, Heidelberg, 2006, pp. 404--417.
\newblock \href {https://doi.org/10.1007/11744023_32}
  {\path{doi:10.1007/11744023_32}}.
\newline\urlprefix\url{https://doi.org/10.1007/11744023_32}

\bibitem{Triggs1999BundleA}
B.~Triggs, P.~F. McLauchlan, R.~I. Hartley, A.~W. Fitzgibbon,
  \href{https://doi.org/10.1007/3-540-44480-7_21}{Bundle adjustment --- a
  modern synthesis}, in: B.~Triggs, A.~Zisserman, R.~Szeliski (Eds.), Vision
  Algorithms: Theory and Practice, Springer Berlin Heidelberg, Berlin,
  Heidelberg, 2000, pp. 298--372.
\newblock \href {https://doi.org/10.1007/3-540-44480-7_21}
  {\path{doi:10.1007/3-540-44480-7_21}}.
\newline\urlprefix\url{https://doi.org/10.1007/3-540-44480-7_21}

\bibitem{Mahmoud}
N.~Mahmoud, A.~Hostettler, T.~Collins, L.~Soler, C.~Doignon, J.~Montiel, Slam
  based quasi dense reconstruction for minimally invasive surgery scenes (05
  2017).

\bibitem{Zhao2016TheEA}
Q.~Zhao, T.~Price, S.~Pizer, M.~Niethammer, R.~Alterovitz, J.~Rosenman,
  \href{https://doi.org/10.1007/978-3-319-46720-7_51}{The endoscopogram: A 3d
  model reconstructed from endoscopic video frames}, in: S.~Ourselin,
  L.~Joskowicz, M.~R. Sabuncu, G.~Unal, W.~Wells (Eds.), Medical Image
  Computing and Computer-Assisted Intervention -- MICCAI 2016, Springer
  International Publishing, Cham, 2016, pp. 439--447.
\newblock \href {https://doi.org/10.1007/978-3-319-46720-7_51}
  {\path{doi:10.1007/978-3-319-46720-7_51}}.
\newline\urlprefix\url{https://doi.org/10.1007/978-3-319-46720-7_51}

\bibitem{monodepth2}
C.~Godard, O.~M. Aodha, M.~Firman, G.~Brostow,
  \href{https://doi.org/10.1109/ICCV.2019.00393}{Digging into self-supervised
  monocular depth estimation}, in: 2019 IEEE/CVF International Conference on
  Computer Vision (ICCV), 2019, pp. 3827--3837.
\newblock \href {https://doi.org/10.1109/ICCV.2019.00393}
  {\path{doi:10.1109/ICCV.2019.00393}}.
\newline\urlprefix\url{https://doi.org/10.1109/ICCV.2019.00393}

\bibitem{Klein2007ParallelTA}
G.~Klein, D.~Murray, \href{https://doi.org/10.1109/ISMAR.2007.4538852}{Parallel
  tracking and mapping for small ar workspaces}, in: 2007 6th IEEE and ACM
  International Symposium on Mixed and Augmented Reality, 2007, pp. 225--234.
\newblock \href {https://doi.org/10.1109/ISMAR.2007.4538852}
  {\path{doi:10.1109/ISMAR.2007.4538852}}.
\newline\urlprefix\url{https://doi.org/10.1109/ISMAR.2007.4538852}

\bibitem{cyclegan}
J.-Y. Zhu, T.~Park, P.~Isola, A.~A. Efros, Unpaired image-to-image translation
  using cycle-consistent adversarial networks, in: 2017 IEEE International
  Conference on Computer Vision (ICCV), 2017, pp. 2242--2251.
\newblock \href {https://doi.org/10.1109/ICCV.2017.244}
  {\path{doi:10.1109/ICCV.2017.244}}.

\bibitem{Chen2016InfoGANIR}
X.~Chen, Y.~Duan, R.~Houthooft, J.~Schulman, I.~Sutskever, P.~Abbeel,
  \href{https://proceedings.neurips.cc/paper_files/paper/2016/file/7c9d0b1f96aebd7b5eca8c3edaa19ebb-Paper.pdf}{Infogan:
  Interpretable representation learning by information maximizing generative
  adversarial nets}, in: D.~Lee, M.~Sugiyama, U.~Luxburg, I.~Guyon, R.~Garnett
  (Eds.), Advances in Neural Information Processing Systems, Vol.~29, Curran
  Associates, Inc., 2016.
\newline\urlprefix\url{https://proceedings.neurips.cc/paper_files/paper/2016/file/7c9d0b1f96aebd7b5eca8c3edaa19ebb-Paper.pdf}

\bibitem{Ranftl}
R.~Ranftl, K.~Lasinger, D.~Hafner, V.~Koltun, Towards robust monocular depth
  estimation: Mixing datasets for zero-shot cross-dataset transfer, IEEE
  Transactions on Pattern Analysis and Machine Intelligence PP (2020) 1--1.
\newblock \href {https://doi.org/10.1109/TPAMI.2020.3019967}
  {\path{doi:10.1109/TPAMI.2020.3019967}}.

\bibitem{Curless1996AVM}
B.~Curless, M.~Levoy, A volumetric method for building complex models from
  range images, Proceedings of the 23rd annual conference on Computer graphics
  and interactive techniques (1996).

\bibitem{unet}
O.~Ronneberger, P.~Fischer, T.~Brox, U-net: Convolutional networks for
  biomedical image segmentation, in: N.~Navab, J.~Hornegger, W.~M. Wells, A.~F.
  Frangi (Eds.), Medical Image Computing and Computer-Assisted Intervention --
  MICCAI 2015, Springer International Publishing, Cham, 2015, pp. 234--241.

\bibitem{resnet}
K.~He, X.~Zhang, S.~Ren, J.~Sun, Deep residual learning for image recognition,
  in: 2016 IEEE Conference on Computer Vision and Pattern Recognition (CVPR),
  2016, pp. 770--778.
\newblock \href {https://doi.org/10.1109/CVPR.2016.90}
  {\path{doi:10.1109/CVPR.2016.90}}.

\bibitem{Masoumian2022MonocularDE}
A.~Masoumian, H.~A. Rashwan, J.~Cristiano, M.~S. Asif, D.~Puig,
  \href{https://www.mdpi.com/1424-8220/22/14/5353}{Monocular depth estimation
  using deep learning: A review}, Sensors 22~(14) (2022).
\newblock \href {https://doi.org/10.3390/s22145353}
  {\path{doi:10.3390/s22145353}}.
\newline\urlprefix\url{https://www.mdpi.com/1424-8220/22/14/5353}

\bibitem{Zhou2013DenseSR}
Q.-Y. Zhou, V.~Koltun, Dense scene reconstruction with points of interest, ACM
  Transactions on Graphics 32 (07 2013).
\newblock \href {https://doi.org/10.1145/2461912.2461919}
  {\path{doi:10.1145/2461912.2461919}}.

\bibitem{Han2024DepthAI}
J.~J. Han, A.~Acar, C.~Henry, J.~Y. Wu,
  \href{https://arxiv.org/abs/2401.16600}{Depth anything in medical images: A
  comparative study} (2024).
\newblock \href {http://arxiv.org/abs/2401.16600} {\path{arXiv:2401.16600}}.
\newline\urlprefix\url{https://arxiv.org/abs/2401.16600}

\end{thebibliography}



\end{document}